\documentclass[11pt,a4paper]{article}
\usepackage{times,latexsym}
\usepackage{url}
\usepackage[T1]{fontenc}

\usepackage[acceptedWithA]{tacl2021v1}
\usepackage{tacl2021v1}

\usepackage{xspace,mfirstuc,tabulary}
\usepackage{mdframed}
\usepackage{float}

\usepackage{booktabs}
\usepackage{pifont}      
\usepackage{xcolor}      
\usepackage{array}       
\usepackage{amssymb}

\usepackage{tikz, forest}
\usetikzlibrary{arrows.meta}

\forestset{
  forked edges/.style={
    for tree={
      edge={-},
      parent anchor=east,
      child anchor=west,
      edge path={
        \noexpand\path[\forestoption{edge}]
        (!u.parent anchor) -- +(5pt,0pt) |- (.child anchor)\forestoption{edge label};
      },
    }
  }
}

\usepackage{enumitem}
\usepackage{placeins}

\usepackage{xstring}
\usepackage{seqsplit}

\newcommand{\capitalfirst}[1]{%
    \StrLeft{#1}{1}[\firstletter]%
    \StrGobbleLeft{#1}{1}[\restofword]%
    \MakeUppercase{\firstletter}\restofword%
}

\newcommand{\capitalizehyphenated}[1]{%
    \StrCut{#1}{-}{\firstpart}{\restpart}%
    \capitalfirst{\firstpart}%
    \IfStrEq{\restpart}{}{}{ \capitalizehyphenated{\restpart}}%
}

\usepackage{todonotes}
\usepackage{soul}

\definecolor{TodoColor}{rgb}{1,0.7,0.6}

\usepackage{multirow}

\definecolor{smartglassesblue}{RGB}{59,95,164}
\definecolor{disfluencypurple}{HTML}{A680B8}
\definecolor{nored}{HTML}{CC0000}

\usepackage{xcolor}
\definecolor{pastelblue}{RGB}{210, 230, 250}

\usepackage{comment}

\usepackage{amsmath}
\usepackage[most]{tcolorbox}

\usepackage[capitalise]{cleveref}
\crefname{figure}{Figure}{Figures}
\crefname{table}{Table}{Tables}
\crefname{appendix}{Appendix}{Appendices}

\definecolor{MariaColor}{rgb}{0.6,0.9,0.6}      
\definecolor{MaikeColor}{rgb}{1.0,0.5,0.5}      
\definecolor{VilemColor}{rgb}{0.5,0.75,1.0}     
\definecolor{FabianColor}{rgb}{1.0,1.0,0.5}     
\definecolor{XiangColor}{rgb}{0.67, 0.9, 0.93}     
\definecolor{shiftLarge}{HTML}{fbb4ae}
\definecolor{shiftMed}{HTML}{fed9a6}
\definecolor{shiftSmall}{HTML}{b3cde3}

\definecolor{tabblue}{HTML}{1F77B4}

\usepackage[table]{xcolor}
\usepackage{makecell}

\usepackage{tcolorbox}

\tcbset{
  sharp corners,    
  frame hidden,     
  left=5pt,         
  right=5pt,        
  top=5pt,          
  bottom=5pt,       
  boxsep=0pt,       
  boxrule=0pt,      
  colback=gray!10,
  colframe=gray!10,
  fontupper=\small,  
  fontlower=\small   
}

\definecolor{tabblue}{HTML}{1F77B4}
\definecolor{tabsalmon}{HTML}{EC9374}
\definecolor{tabblue}{HTML}{3986BC}

\newtoggle{short}

\newcommand{\shortlong}[2]{%
  \iftoggle{short}{#1}{#2}%
}

\newcommand{\hlc}[2][yellow]{{%
    \colorlet{foo}{#1}%
    \sethlcolor{foo}\hl{#2}}%
}

\definecolor{hlCap}{RGB}{255, 235, 130}
\definecolor{btnPurple}{RGB}{108, 92, 125}
\definecolor{btnGray}{RGB}{238, 233, 242}

\newif\iftaclinstructions
\taclinstructionsfalse 
\iftaclinstructions
\renewcommand{\confidential}{}
\renewcommand{\anonsubtext}{(No author info supplied here, for consistency with
TACL-submission anonymization requirements)}
\newcommand{\instr}
\fi

\iftaclpubformat 

\else

\fi

\toggletrue{short}

\title{The Role of Disfluencies in Speech Translation}

\author{
  \textbf{Maike Züfle$^{1}$}\hspace{-4pt}
  \Thanks{Equal contribution.}
  \quad
  \textbf{Maria Teleki$^{2*}$}
  \quad
  \textbf{Fabian Retkowski$^{1}$}
  \quad
  \textbf{Vilém Zouhar$^{3}$}
  \\
  \textbf{Oliver Grabner$^{2}$}
  \quad
  \textbf{Alexander Waibel$^{1,4}$}
  \quad
  \textbf{James Caverlee$^{2}$}
  \quad
  \textbf{Jan Niehues$^{1}$}
  \\
  \ \\
  $^1$Karlsruhe Institute of Technology \quad
  $^2$Texas A\&M University \\
  $^3$ETH Zurich \quad
  $^4$Carnegie Mellon University
  \\ \ \\
  \texttt{maike.zuefle@kit.edu}
}

\date{}

\begin{document}
\maketitle
\begin{abstract}
    Current speech translation systems, including SpeechLLMs, are trained on cleaned text and tend to strip disfluencies like filled pauses and false starts rather than translate them. We show this comes at a cost: disfluencies carry meaning that gets lost when speech is cleaned up. To study this systematically, we introduce \textsc{Uh-Mazing}, a benchmark of human-translated, disfluency-annotated Switchboard speech covering English into eight target languages. Across these languages and several architectures, we find that false starts and self-repairs, not filled pauses or discourse markers, drive most of the translation-quality loss, and that models which fail to preserve a disfluency tend to omit it rather than mistranslate it.  We show inference-time decoding can mitigate this without retraining, and release the benchmark and code.
\end{abstract}
\section{Introduction}

Spontaneous speech is disfluent \cite{Shriberg_1994}. Filled pauses, repetitions, false starts, and self-repairs occur frequently in natural conversation and reflect underlying cognitive and discourse processes. While such phenomena are visible to human listeners, they pose persistent challenges for automatic speech processing systems \cite{liesenfeld-etal-2023-timing, teleki24_interspeech}. Faced with such limitations, conversational participants often adapt their own speech to cope with live system shortcomings \cite{hara2015effect, akira2017speech}, underscoring how disfluency handling shapes real-world interaction even before translation quality is measured.

Speech translation is increasingly deployed in settings where disfluencies carry information that must not be lost \citep{rice2025mashti}: in assistive wearables, where a speaker's hesitation or self-correction can be as informative as their words \citep{metaglasses, transcribeglass, xanderglasses}; in multilingual meetings, where filled pauses and false starts signal turn-taking and uncertainty \citep{meetween2024}, and in  dubbing, where stripping disfluencies can make a speaker sound more confident or fluent than they actually were. Not every setting calls for this, however: live captioning and simultaneous interpretation often favour a fluent rendering instead \citep{papi-etal-2023-direct-speech, papi-etal-2025-real}, since disfluencies add reading or listening effort under time pressure without a compensating benefit \citep{choi26_speechspectrum}. 

Yet across the many settings where preserving disfluencies does matter, most speech translation systems still treat disfluency as noise: end-to-end models remove them explicitly \citep{salesky-etal-2019-fluent, salesky-et-al-2018-disfluent, dong-disfluent-2019}, cascaded systems in which ASR omits disfluencies before they reach the translation step \citep{jamshid-lou-johnson-2020-end, amann-asr-2024, kumar-etal-2026-mind}, and contemporary LLMs are trained predominantly on cleaned text \citep{retkowski-etal-2025-summarizing}. Prior work shows this comes at a cost: disfluencies encode affective content, interpersonal relationships, and discourse structure \citep{campbell2007translating, kundu2022survey}, and their removal reduces perceived naturalness \citep{hassan2025enhancingnaturalnessllmgeneratedutterances}.
Recent SpeechLLMs~\citep{arora2025landscapespokenlanguagemodels} reduce reliance on 
intermediate text representations by operating directly on speech signals, but evidence 
suggests they face the same blind spot~\citep{teleki25_dres, liu2025vocalbench}: existing 
benchmarks largely focus on fluent speech or implicitly reward disfluency removal
~\citep{teleki-etal-2024-quantifying-impact, choi26_speechspectrum}, providing limited 
insight into model behaviour under realistic, conversational conditions. Where disfluency-preserving translation data exists at all, it is limited to a single language pair \citep{post-etal-2013-improved}, and the translation is not designed or annotated for disfluency research. \shortlong{}{We present a more detailed discussion of related work in \cref{app:rel_work}.}

In this work, we first ask why disfluencies matter: we show in a human case study that removing disfluencies from speech shifts perceived negative emotion, by as much as 71 percentage points in the most extreme case, motivating their preservation. Building on this, we introduce \textsc{Uh-Mazing}, a disfluency-aware speech translation benchmark of carefully curated human-translated Switchboard speech spanning eight target languages, with fine-grained  annotations of disfluency type.

Using this benchmark, we analyse how current speech translation models, including SpeechLLMs, handle disfluencies. We find that false starts and self-repairs, not filled pauses or discourse markers, are primarily responsible for the translation-quality cost of disfluent speech. Through human annotation of model outputs into error categories covering both general translation errors and disfluency-specific errors, we show that when models fail to preserve a disfluency, they typically omit it outright rather than mistranslating it. Beyond this analysis, we explore inference-time decoding strategies that improve disfluency preservation without retraining, providing practical guidance for deploying these systems in real-world, conversational settings.
We contribute:
\begin{itemize}[left=0mm]
    \item A human evaluation showing that disfluencies carry emotionally relevant information.

    \item \textsc{Uh-Mazing}, a human-annotated disfluency-aware speech translation benchmark with eight target languages and disfluency annotations.

    \item A human-annotated taxonomy of disfluency-specific failure modes in current speech translation systems.

    \item A systematic evaluation of inference-time strategies, showing that beam search and in-context learning offer simple, low-risk improvements to disfluency handling.
\end{itemize}

\noindent
 Since the \textsc{Uh-Mazing} dataset is built on Switchboard \citep{godfrey1993switchboard}, we will submit it to \href{https://www.ldc.upenn.edu/data-management/providing-data}{Penn LDC} for public release.\footnote{Experimental code and the \textsc{Uh-Mazing} dataset will be released upon LDC approval.}

\shortlong{}{\begin{figure}[t]
    \centering
    \includegraphics[width=0.95\linewidth]{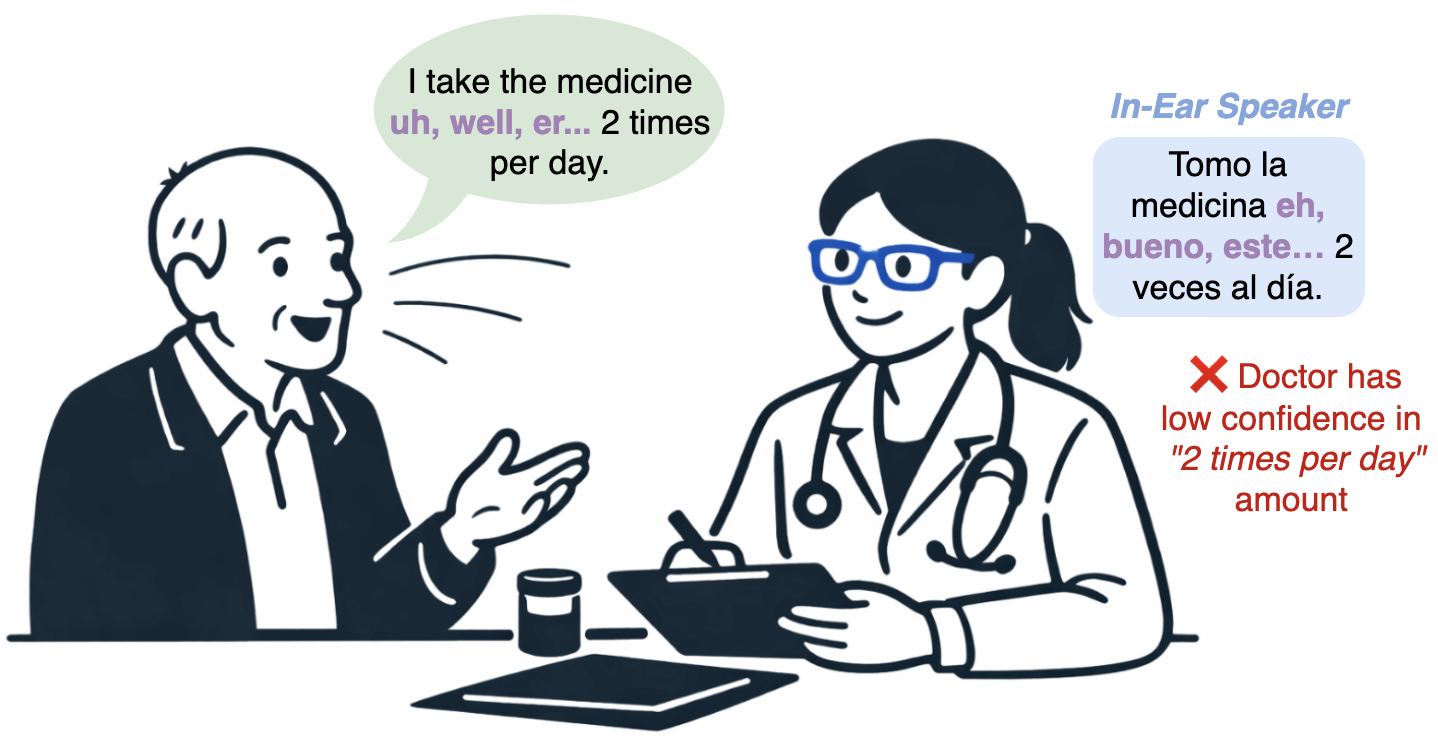}
    \caption{\textbf{Uncertainty lives in the disfluencies; translations that strip them strips the uncertainty with them.} In this translation with \textcolor{smartglassesblue}{\textbf{smart glasses}}, the \textcolor{disfluencypurple}{\textbf{disfluencies}} impact the doctor's \textcolor{nored}{\textbf{confidence assessment}}.}
    \label{fig:intro}
\end{figure}
}

\section{The Role of Disfluencies}
\label{sec:role_of_disfluencies}

Disfluencies are natural features of spontaneous speech, yet NLP pipelines often treat them as noise and remove them~\cite{teleki25_icwsm, choi2026singlegroundtruthreference}. In this section, we investigate what disfluencies communicate and argue that they carry emotionally relevant information that should be preserved, particularly in speech translation.

\subsection{Methodology}\label{subsec:meaning-methodology}
\paragraph{Data Selection: Switchboard.}
We use the Switchboard corpus \citep{godfrey1993switchboard}, which consists of spontaneous English conversational telephone speech with detailed disfluency annotations, making it well-suited for studying genuine disfluent speech \citep{kundu2022survey}. We align the transcripts with the corresponding audio recordings and retain the original disfluency annotations. Following \citet{Shriberg_1994}, we focus on three disfluency classes: interjections (\textsc{intj}), parentheticals (\textsc{prn}), and edited speech (\textsc{edited}).

\paragraph{Human annotation of disfluency meaning.}
We select 80 utterances from the Switchboard corpus, manually chosen to contain multiple disfluencies and to span a range of lengths (17--212 words, median~51; totalling 25 minutes of audio). To assess how disfluencies shape perceived emotion, we ask human raters to rate the emotional content of each utterance under three conditions: \textit{Audio}, which carries both disfluencies and paralinguistic cues (tone, prosody, timing); \textit{Disfluent}, which preserves disfluencies in text but strips paralinguistic cues; and \textit{Fluent}, which strips both.

We recruit 18 annotators and ask each to evaluate 40 utterances under one condition per utterance using the Pearmut platform \citep{zouhar2026pearmuthumanevaluationtranslation}. Each utterance is rated by three annotators per condition, with different annotators assigned to each condition, so no annotator sees the same utterance in more than one form.
 Annotators can annotate up to five free-text emotions and rate each on a 0--5 intensity scale (interface in \cref{fig:role_of_disfluency_pearmut}).

In addition, we also analyse whether LLMs can reproduce these human emotion judgments. To test this, we also replicate the evaluation with \texttt{GPT-5.2-2025-12-11} across seven seeds. \shortlong{}{More details can be found in \cref{app:llm-replication}.}

\paragraph{Analysis approach.} 
Open-ended emotion labels are difficult to compare across annotators, so we structure the analysis around three principles: 

First, to ensure findings do not depend on the categorisation method, we use two independent pipelines. The first is NRC EmoLex \citep{mohammad2013crowdsourcing}, a lexicon mapping free-form labels (e.g., \emph{worried}, \emph{frustrated}, \emph{hopeful}) to eight basic emotions \citep{plutchik1980general}, grouped by valence into negative (anger, fear, sadness, disgust), positive (joy, trust), and variable (surprise, anticipation). Because NRC EmoLex indexes surface forms only, we map each label via a three-pass cascade: direct surface match, Snowball-stem lookup \citep{Porter2001SnowballAL}, and WordNet-lemma lookup \citep{miller-1994-wordnet}. \shortlong{}{Full validation can be found in \cref{app:nrc-validation}. }The second categorisation method is an OpenAI-embedding pipeline, which embeds each free-form label with \texttt{text-embedding-3-small} and assigns it a valence (negative, neutral, or positive) by nearest cosine similarity to seed words. 

Second, we also analyse the magnitude of emotional shifts and not only their existence. Hence, we report Cramér's $V$ and $\varepsilon^2$ rather than $p$-values. 

Third, to verify that no single annotator or sample drives the results, we rerun the analysis leaving out each annotator in turn (18 iterations) and each sample in turn (80 iterations), and report only findings that survive both with tight effect-size ranges.

\begin{figure}[t]
  \centering
  \includegraphics[width=\linewidth]{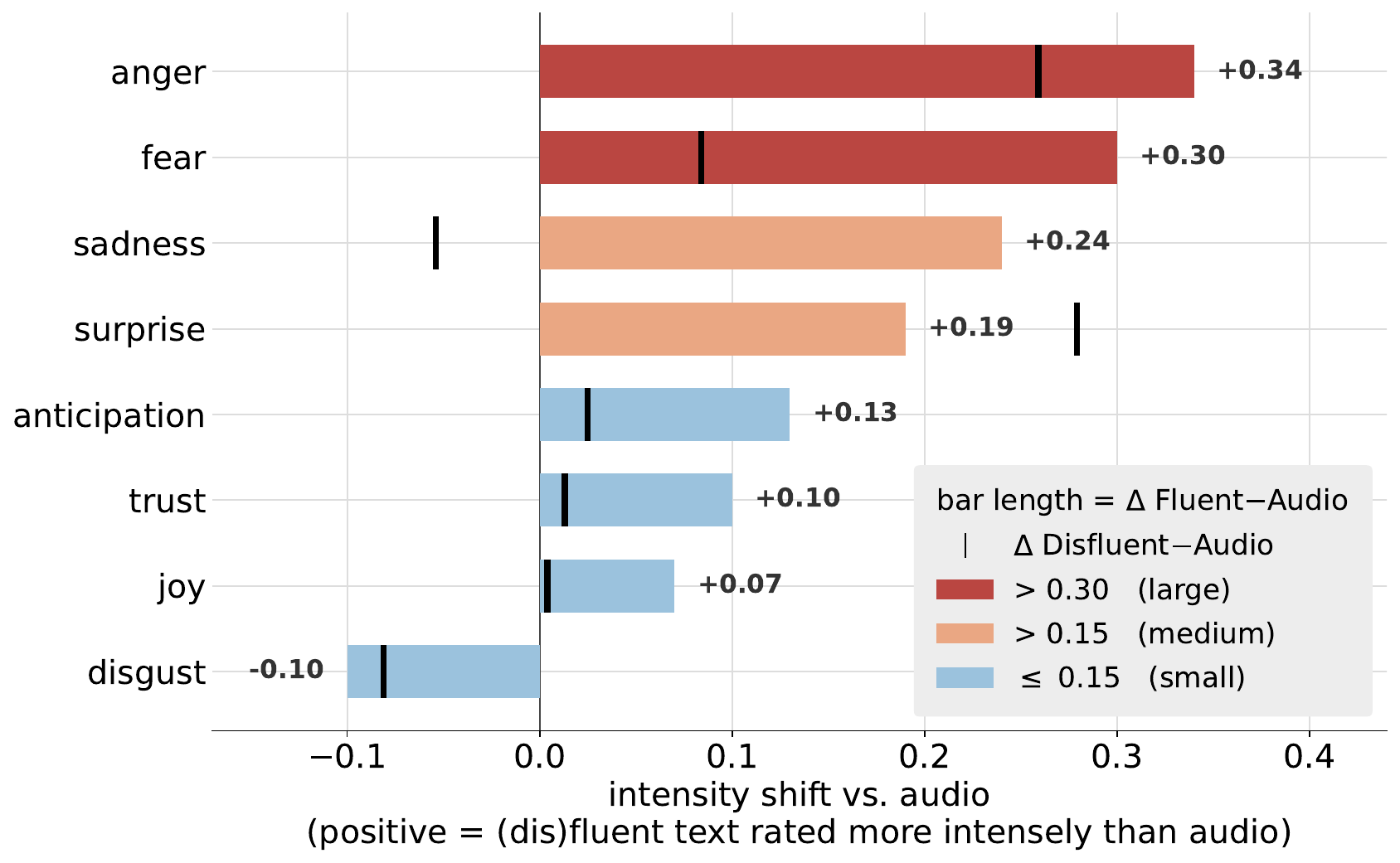}
  \caption{\textbf{Removing audio signal and disfluencies increases perceived intensity for seven of the eight emotions. Disgust reverses.} Six of the eight increase at both steps: Audio$\to$Disfluent and Disfluent$\to$Fluent. A tick inside the bar marks these emotions.}
  \label{fig:nrc-slopes}
\end{figure}

\subsection{Analysis of Disfluency Annotations}

Annotators produced 562 distinct free-form emotion labels, of which 66\% map to NRC EmoLex categories. The remaining 34\% are covered by the embedding pipeline. The most frequently annotated emotions are trust ($n{=}782$), sadness ($n{=}657$), and anticipation ($n{=}558$), while surprise is the least common ($n{=}288$). Per-condition intensity breakdowns for all eight emotions are shown in \cref{fig:intensity_emotions} in the appendix.

\paragraph{Removing disfluencies amplifies perceived negative emotion.}
Fluent text generally carries the strongest negative emotional readings, adding disfluencies softens these, and hearing the original audio softens them further (\cref{fig:nrc-slopes}). Two emotions deviate from this pattern: disgust reverses entirely, with audio carrying the strongest reading, while surprise peaks at the disfluent condition. The two steps contribute differently: removing paralinguistic signal (Audio $\to$ Disfluent) accounts for the largest shift in anger ($+0.23$) and surprise ($+0.28$), while removing disfluencies (Disfluent $\to$ Fluent) drives the larger shifts in fear ($+0.21$) and sadness ($+0.29$). \shortlong{}{Full per-condition means are in \cref{tab:nrc-dir}.}

The embedding pipeline, which assigns each label one of three valence categories (negative, neutral, positive), shows the same overall direction. Both text conditions carry more negative labels than audio (Audio 30\%, Disfluent 39\%, Fluent 34\%, $V{=}0.07$), and the neutral share is lowest for the disfluent transcript (Audio 34\%, Disfluent 25\%, Fluent 31\%, $V{=}0.08$). Within the text conditions, the disfluent transcript produces the highest share of negative labels, while negative-label intensity is highest for fluent text (Audio 2.49, Disfluent 2.68, Fluent 2.78, $\varepsilon^2{=}0.01$). We show qualitative examples in \cref{tab:qual-examples} in the appendix.

\paragraph{Annotators diverge more when disfluencies are present.}
\label{sec:iaa}
We next examine whether annotators agree more on some conditions than others. For each item annotated by at least two workers, we compute the fraction agreeing on the dominant valence category. Fluent transcripts show the highest agreement (67.3\%), while Audio (60.5\%) and Disfluent (60.6\%) are significantly lower (Wilcoxon signed-rank, both $p<0.01$). When disfluencies are present, whether in audio or the transcript, listeners diverge more in what they think the speaker is feeling. This confirms that disfluencies are not noise: their presence shifts how listeners interpret the speaker's emotional state, consistent with the view of disfluencies as interpretive hedges~\citep{brennan2000invited,diachek2024linguistic}.

\paragraph{Can LLMs replace human annotators?}
The interpretive complexity established above raises the question of whether LLMs could substitute for human annotators at scale. While LLM seeds are highly self-consistent ($r{=}0.86$ between seeds), human-to-LLM correlation averages only $r{=}0.35$ across all seven seeds, well below the human-human ceiling of $r{=}0.54$. This gap does not shrink under seed variation, indicating that LLMs cannot yet substitute for humans in this task. \shortlong{}{(Full details and results can be found in \cref{app:llm-replication})}

\paragraph{Implications for translation.}
The findings above paint a consistent picture. Removing disfluencies amplifies perceived negative emotion, and listeners diverge more on disfluent content than on fluent text. Disfluencies should therefore be preserved in translation: they carry interpretive signal that belongs to the receiver to resolve, not the translator~\citep{angelelli2004medical}.
Our benchmark accordingly evaluates whether translation systems preserve disfluencies in the target language.

\section{\textsc{Uh-Mazing} Speech Translation}
\label{sec:Uh-Mazing Speech Translation}
Building on the Switchboard utterances selected in \cref{sec:role_of_disfluencies}, we construct \textsc{Uh-Mazing}, a multilingual benchmark for disfluency-aware speech translation. We translate the samples into eight target languages, allowing us to directly examine how disfluencies are handled across typologically diverse languages and to evaluate automatic translation systems in this regard.

\subsection{Human Translation} \label{sec:human-translation}

\textsc{Uh-Mazing} consists of 640 human translations of 80 carefully curated Switchboard utterances~\citep{godfrey1993switchboard}, which were selected for high disfluency density and a wide range of lengths (\cref{sec:role_of_disfluencies}). Each utterance is translated from English into Mandarin Chinese (ZH), Spanish (ES), Hindi (HI), French (FR), German (DE), Italian (IT), Czech (CS), and Modern Standard Arabic (AR). Target languages were chosen to span diverse language families and regions, enabling analysis of how disfluency and translation behaviour vary across languages.

For each pair, the source side consists of the speech recording, its time-aligned disfluent transcript, and a fluent transcript obtained by stripping the marked disfluencies (edited, interjections, parentheticals); the target side consists of a disfluent and a fluent translation, produced by the annotation process described below.

\paragraph{Annotation Protocol.}
Translations were collected via Prolific in two rounds with independently recruited annotators, following \citet{taguchi2025languages}. In \textit{Round 1}, annotators translated each disfluent English source into the target language in batches of 20 utterances, marking disfluent tokens with underscores (e.g.\ \textit{\_uh\_}). In \textit{Round 2}, a second annotator per language reviewed each Round 1 translation. This standardises the disfluency markup and provides a per-translation confidence signal, since agreement between the two rounds is stronger evidence of correctness than either round alone. We use Round 2 outputs as the final dataset. \shortlong{}{Annotator counts, compensation, and bonus payments are in \cref{sec:humeval_guidelines}.} The annotation quality of Prolific participants has been validated in prior work~\cite{gordon2026aiagent}, and we further validate it against our expert translators below.

\subsection{Inter-annotator agreement on translation}
\label{sec:translation-iaa}

To quantify translator agreement on disfluent speech, we compute inter-annotator agreement on ten paired translations per target language across all eight languages. For each item, one translation comes from the Prolific crowdworker cohort described in \cref{sec:human-translation}. The paired reference translation was produced by an author or a native-speaker collaborator serving as an expert annotator. We embed each translation with OpenAI's \texttt{text-embedding-3-large} after removing the underscore-delimited disfluency markup, and report cosine similarity between the two embeddings. Overall agreement is high: mean expert$\leftrightarrow$Prolific cosine is $0.878$ ($\sigma{=}0.075$), and 71 of 80 pairs sit above $0.80$. The ordering is stable across three different OpenAI embedders (\cref{fig:translation-iaa}). The Prolific translations therefore reliably track expert reference translations, and we use them as our translation data in this work.

\subsection{The Role of Disfluencies in Different Languages}

\begin{figure}
    \centering
    \includegraphics[width=1.0\linewidth]{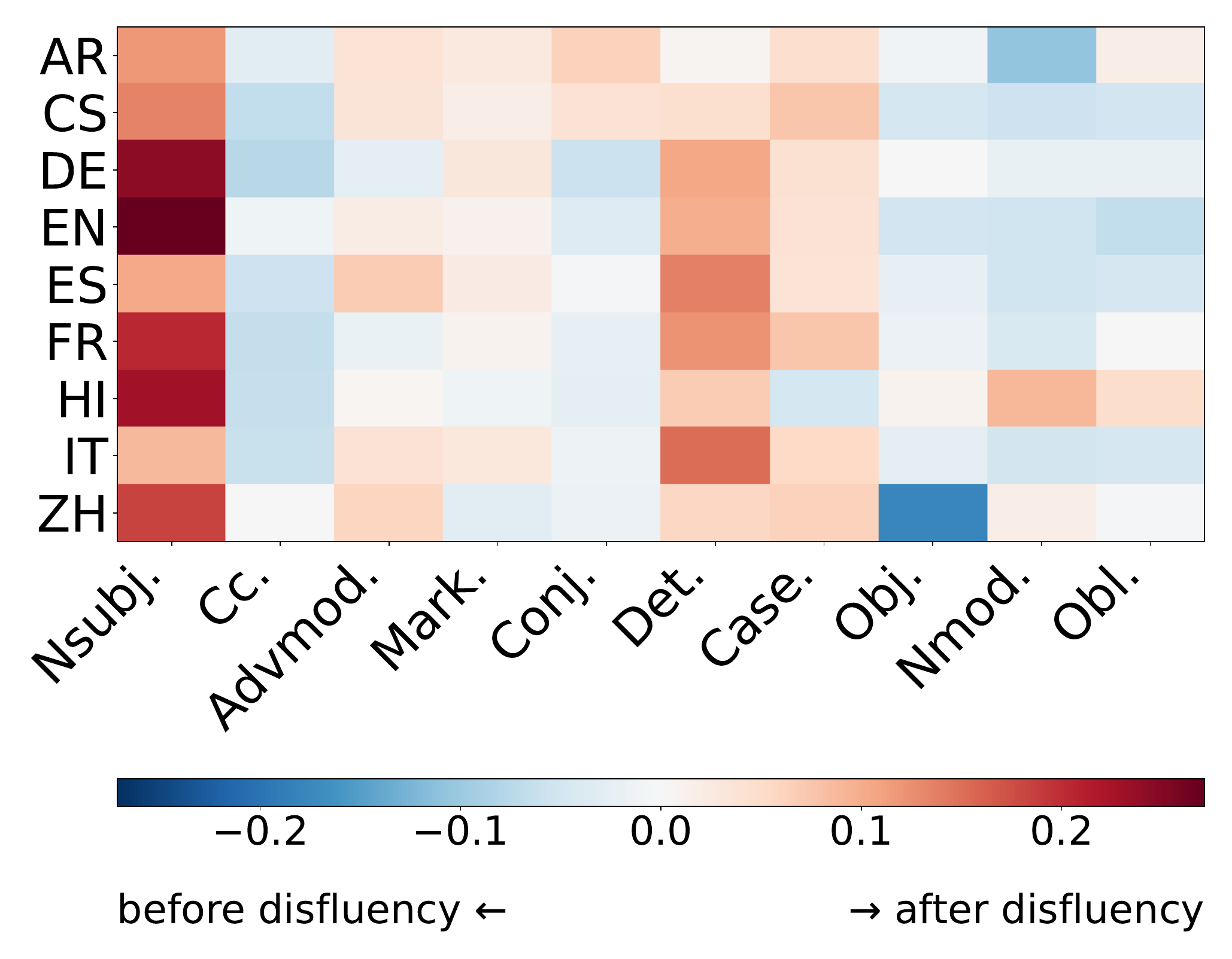}
    \caption{\textbf{Across languages, disfluencies most often follow a coordinating conjunction and precede the sentence subject.} Position of disfluencies by the dependency relation of neighboring content words.}
    \label{fig:disfluency_pos}
\end{figure}

We now analyse how the disfluencies are realised across different languages. While disfluency counts (428--482 per language) are broadly comparable across languages (\cref{tab:basic-statistics} in the appendix),  average disfluency length varies from 2.7 words (Arabic) to 3.8 words (Chinese). 

\paragraph{Syntactic Context of Disfluencies.}
We parse the reference translations for all eight target languages with the Stanza dependency parser~\citep{qi-etal-2020-stanza}, characterising each disfluency by the dependency relation of the nearest fluent content word immediately before and after it (\cref{fig:disfluency_pos}).

Across nearly all eight languages, disfluencies systematically precede the sentence's subject rather than follow it, a pattern that holds with consistency across typologically diverse languages. The word preceding a disfluency is instead most often a coordinating conjunction (\textsc{cc}), consistent with disfluencies frequently arising
right after connectives such as \emph{and} or \emph{but}. The strength of this asymmetry still differs by language.

\paragraph{Context Shift on Translation.}
We next ask whether the syntactic context of a disfluency is preserved when the utterance is translated into another language\shortlong{}{(\Cref{tab:context-shift-by-category} in the appendix)}. It is not: the contextual placement varies substantially by language and disfluency category. Markers introducing subordinate clauses retain their context comparatively more often (9--42\% across languages), whereas discourse markers do so in under 8\% of cases in every language. This shows that when a disfluency is translated, its syntactic environment is rarely carried over unchanged, another source of the difficulty translation systems face with disfluent speech.

\section{Models \& Evaluation}
\label{sec:eval}
Having established that disfluencies carry emotionally relevant information, we now turn to the question of whether current speech translation models are capable of preserving them.

\subsection{Translation Models}

We evaluate four model types (cascaded, end-to-end, SpeechLLM, and commercial), using both a standard and a disfluency-aware prompt for all except end-to-end models, which are not designed to follow explicit instructions (prompts in \Cref{fig:prompts}).

\begin{itemize}[left=0mm] 
\item \textbf{Cascaded:} Canary~\cite{sekoyan2025canary1bv2parakeettdt06bv3efficient} or Whisper~\cite{radford2022whisper} for ASR, followed by Tower~\cite{rei2025towerplus} or Llama~\cite{grattafiori2024llama3herdmodels} for translation; also evaluated on human-annotated transcripts.
\item \textbf{End-to-end:} Canary and OWSM~\cite{owsm}. OWSM: does not support Hindi; Canary: does not support Chinese, Hindi, or Arabic.
\item \textbf{SpeechLLM:} Phi-4~\cite{abdin2024phi4technicalreport} and Qwen2.5-Omni~\cite{xu2025qwen25omnitechnicalreport}.
\item \textbf{Commercial:} ChatGPT (GPT-5.2-2025-12-11) and Gemini ~\cite[gemini-2.5-flash]{comanici2025gemini25pushingfrontier}, evaluated with both speech input and human-annotated transcripts.
\end{itemize}

\subsection{Evaluation}\label{sec:eval_methods}
We evaluate all model families in terms of overall translation quality and disfluency preservation.

\paragraph{Automatic Metrics.} We assess translation quality using chrF \cite[word-segmenting Chinese translations using \texttt{jieba}]{popovic-2015-chrf}, which measures character-level overlap with the reference and is therefore sensitive to disfluency preservation, and COMET-Kiwi \cite{rei-etal-2022-comet}, a reference-free metric that focuses on semantic adequacy. For cascaded models, we additionally report $\mathcal{E}$-Scores and $\mathcal{Z}$-Scores \cite{teleki25_zscores} at the ASR step, which measure the retention of disfluent elements.

\paragraph{Disfluency Preservation.} 
To evaluate whether models correctly preserve disfluencies in the translation, we additionally use LLM-as-a-judge \cite{zheng2023judging, li2025generation} with \texttt{GPT-5.2-2025-12-11} to score translations on two dimensions (0--100):
\begin{itemize}[left=0mm]
    \item \textbf{Style Preservation:} How well are disfluencies preserved (presence, type, placement, feel)?
    \item \textbf{Meaning Preservation:} How accurately and completely is the source meaning preserved?
\end{itemize}
We use two prompt variants for the LLM judge: with and without a reference translation (full prompts in \cref{fig:prompts}\shortlong{}{ in \cref{app:llm-judge}}).  We select the best-performing model and prompting strategy from each category based on automatic metrics, and apply both LLM judge variants to these systems.

\paragraph{Human Evaluation of Quality.}
Finally, we complement automatic metrics with human evaluation, inspired by MQM and ESA annotation protocols \citep{freitag-etal-2021-experts,kocmi-etal-2024-error}. We select the same best-performing models as for the LLM judge, and present their outputs side by side.  Each item is annotated by one annotator, with the first 20 items additionally annotated by two further annotators to confirm inter-annotator reliability (leave-one-out soft pairwise accuracy: 96.1\% for style, 94.0\% for meaning). Annotators score each translation on the same dimensions as the LLM judge, Style Preservation and Meaning Preservation, but additionally mark and categorise error spans. We use the Pearmut platform \citep{zouhar2026pearmuthumanevaluationtranslation}. Human annotation guidelines and the interface are shown in \Cref{fig:pearmut_quality}.

\begin{table}[t]
\centering
\begin{tabular}{llrr}
\toprule
\textbf{Metric} & \textbf{\shortstack{Ref-\\based?}} & \textbf{\shortstack{Segm.\\Corr.}} & \textbf{\shortstack{Sys.\\Corr.}} \\
\midrule
\multicolumn{4}{c}{\textit{Style}} \\
\midrule
\mbox{LLM-judge}\textsubscript{Style} & $\times$ & 37.3 & 86.8 \\
\mbox{LLM-judge}\textsubscript{Style} & $\checkmark$ & \textbf{40.4} & 92.5 \\
chrF & $\checkmark$ & 38.2 & \textbf{93.1} \\
\mbox{COMET-Kiwi} & $\times$ & 26.9 & 78.2 \\
\midrule
\multicolumn{4}{c}{\textit{Meaning}} \\
\midrule
\mbox{LLM-judge}\textsubscript{Meaning} & $\times$ & 34.2 & \textbf{90.2} \\
\mbox{LLM-judge}\textsubscript{Meaning} & $\checkmark$ & \textbf{40.8} & \textbf{90.2} \\
chrF & $\checkmark$ & 34.6 & 88.4 \\
\mbox{COMET-Kiwi} & $\times$ & 33.4 & 83.7 \\
\midrule
\multicolumn{4}{c}{\textit{Overall}} \\
\midrule
\mbox{LLM-judge}\textsubscript{Overall} & $\times$ & 42.1 & 91.0 \\
\mbox{LLM-judge}\textsubscript{Overall} & $\checkmark$ & \textbf{46.8} & 92.6 \\
chrF & $\checkmark$ & 38.6 & \textbf{92.8} \\
\mbox{COMET-Kiwi} & $\times$ & 30.9 & 80.4 \\
\bottomrule
\end{tabular}
\caption{\textbf{Reference-based metrics have best correlation with human judgments on disfluent translations.} Correlation (\%) between human Style/Meaning/Overall judgments and automatic metrics, averaged over the eight languages and four systems. Segment-level: mean Kendall's $\tau$ (within-item, across systems). System-level: soft pairwise accuracy (SPA).}
\label{tab:correlation}
\end{table}
\section{Results}
\label{sec:Results}
\begin{table*}[t]
\centering
\resizebox{\textwidth}{!}{%
\begin{tabular}{llrrrrrrrr}
\toprule
\textbf{Category} & \textbf{Model} & \textbf{AR} & \textbf{CS} & \textbf{DE} & \textbf{ES} & \textbf{FR} & \textbf{HI} & \textbf{IT} & \textbf{ZH} \\
\midrule
\multicolumn{10}{c}{\textit{EN $\to$ X: Speech-to-Text Translation}} \\
\midrule
\multirow{6}{*}{Cascaded} & \textsc{Llama}\textsubscript{Canary ASR} & \cellcolor{tabsalmon!0}{22.16} & \cellcolor{tabsalmon!1}{38.37} & \cellcolor{tabsalmon!2}{49.30} & \cellcolor{tabsalmon!3}{52.53} & \cellcolor{tabsalmon!1}{52.12} & \cellcolor{tabsalmon!7}{35.91} & \cellcolor{tabsalmon!1}{50.06} & \cellcolor{tabsalmon!0}{13.42} \\
 & \textsc{Llama}\textsubscript{Whisper ASR} & \cellcolor{tabsalmon!17}{30.02} & \cellcolor{tabsalmon!0}{37.96} & \cellcolor{tabsalmon!0}{48.19} & \cellcolor{tabsalmon!0}{51.78} & \cellcolor{tabsalmon!0}{51.69} & \cellcolor{tabsalmon!12}{38.99} & \cellcolor{tabsalmon!0}{49.11} & \cellcolor{tabsalmon!0}{13.59} \\
\addlinespace[3pt]
 & \textsc{Llama}\textsubscript{Human transcript} & \cellcolor{tabsalmon!0}{16.64} & \cellcolor{tabsalmon!18}{43.97} & \cellcolor{tabsalmon!28}{56.30} & \cellcolor{tabsalmon!39}{61.40} & \cellcolor{tabsalmon!31}{59.29} & \cellcolor{tabsalmon!3}{33.25} & \cellcolor{tabsalmon!35}{59.11} & \cellcolor{tabsalmon!13}{17.07} \\
\addlinespace[3pt]
 & \textsc{Tower}\textsubscript{Canary ASR} & \cellcolor{tabsalmon!1}{22.87} & \cellcolor{tabsalmon!20}{44.43} & \cellcolor{tabsalmon!19}{53.86} & \cellcolor{tabsalmon!11}{54.40} & \cellcolor{tabsalmon!16}{55.67} & \cellcolor{tabsalmon!35}{52.24} & \cellcolor{tabsalmon!13}{53.21} & \cellcolor{tabsalmon!29}{21.13} \\
 & \textsc{Tower}\textsubscript{Whisper ASR} & \cellcolor{tabsalmon!3}{23.42} & \cellcolor{tabsalmon!18}{43.90} & \cellcolor{tabsalmon!11}{51.76} & \cellcolor{tabsalmon!7}{53.40} & \cellcolor{tabsalmon!5}{53.03} & \cellcolor{tabsalmon!34}{51.93} & \cellcolor{tabsalmon!8}{51.87} & \cellcolor{tabsalmon!27}{20.68} \\
\addlinespace[3pt]
 & \textsc{Tower}\textsubscript{Human transcript} & \cellcolor{tabsalmon!0}{22.39} & \cellcolor{tabsalmon!46}{52.90} & \cellcolor{tabsalmon!47}{61.34} & \cellcolor{tabsalmon!50}{\textbf{64.05}} & \cellcolor{tabsalmon!50}{\textbf{63.65}} & \cellcolor{tabsalmon!50}{\textbf{61.43}} & \cellcolor{tabsalmon!50}{\textbf{62.93}} & \cellcolor{tabsalmon!48}{25.89} \\
\midrule
\multirow{2}{*}{End-to-end} & \textsc{OWSM} & \cellcolor{tabsalmon!9}{26.47} & \cellcolor{tabsalmon!0}{27.38} & \cellcolor{tabsalmon!0}{47.76} & \cellcolor{tabsalmon!0}{43.02} & \cellcolor{tabsalmon!0}{43.17} & -- & \cellcolor{tabsalmon!0}{40.93} & \cellcolor{tabsalmon!6}{15.37} \\
 & \textsc{Canary} & -- & \cellcolor{tabsalmon!18}{43.89} & \cellcolor{tabsalmon!13}{52.13} & \cellcolor{tabsalmon!0}{51.65} & \cellcolor{tabsalmon!10}{54.30} & -- & \cellcolor{tabsalmon!11}{52.80} & -- \\
\midrule
\multirow{2}{*}{SpeechLLM} & \textsc{Phi-4-MM-instruct} & \cellcolor{tabsalmon!0}{5.73} & \cellcolor{tabsalmon!0}{8.01} & \cellcolor{tabsalmon!0}{45.79} & \cellcolor{tabsalmon!0}{46.03} & \cellcolor{tabsalmon!0}{27.12} & \cellcolor{tabsalmon!0}{3.56} & \cellcolor{tabsalmon!0}{49.58} & \cellcolor{tabsalmon!2}{14.28} \\
 & \textsc{Qwen2.5-Omni} & \cellcolor{tabsalmon!0}{17.32} & \cellcolor{tabsalmon!0}{10.61} & \cellcolor{tabsalmon!0}{20.90} & \cellcolor{tabsalmon!0}{18.02} & \cellcolor{tabsalmon!0}{18.27} & \cellcolor{tabsalmon!0}{11.78} & \cellcolor{tabsalmon!0}{21.28} & \cellcolor{tabsalmon!0}{12.65} \\
\midrule
\multirow{6}{*}{Commercial} & \textsc{Gemini}\textsubscript{ASR transcript} & \cellcolor{tabsalmon!40}{40.51} & \cellcolor{tabsalmon!47}{53.27} & \cellcolor{tabsalmon!45}{60.91} & \cellcolor{tabsalmon!35}{60.46} & \cellcolor{tabsalmon!47}{62.95} & \cellcolor{tabsalmon!38}{54.30} & \cellcolor{tabsalmon!46}{61.91} & \cellcolor{tabsalmon!40}{23.79} \\
\addlinespace[3pt]
 & \textsc{Gemini}\textsubscript{Human transcript} & \cellcolor{tabsalmon!35}{38.00} & \cellcolor{tabsalmon!25}{46.18} & \cellcolor{tabsalmon!14}{52.62} & \cellcolor{tabsalmon!3}{52.51} & \cellcolor{tabsalmon!7}{53.44} & \cellcolor{tabsalmon!26}{47.18} & \cellcolor{tabsalmon!14}{53.60} & \cellcolor{tabsalmon!24}{19.87} \\
\addlinespace[3pt]
 & \textsc{Gemini}\textsubscript{Audio} & \cellcolor{tabsalmon!33}{37.15} & \cellcolor{tabsalmon!34}{48.97} & \cellcolor{tabsalmon!24}{55.24} & \cellcolor{tabsalmon!28}{58.54} & \cellcolor{tabsalmon!21}{56.83} & \cellcolor{tabsalmon!0}{15.90} & \cellcolor{tabsalmon!30}{57.60} & \cellcolor{tabsalmon!0}{7.13} \\
\addlinespace[3pt]
 & \textsc{ChatGPT}\textsubscript{ASR transcript} & \cellcolor{tabsalmon!50}{\textbf{44.97}} & \cellcolor{tabsalmon!50}{\textbf{54.19}} & \cellcolor{tabsalmon!50}{\textbf{62.28}} & \cellcolor{tabsalmon!37}{60.92} & \cellcolor{tabsalmon!43}{62.12} & \cellcolor{tabsalmon!41}{56.17} & \cellcolor{tabsalmon!40}{60.34} & \cellcolor{tabsalmon!50}{\textbf{26.42}} \\
\addlinespace[3pt]
 & \textsc{ChatGPT}\textsubscript{Human transcript} & \cellcolor{tabsalmon!43}{41.74} & \cellcolor{tabsalmon!24}{45.82} & \cellcolor{tabsalmon!19}{53.97} & \cellcolor{tabsalmon!7}{53.35} & \cellcolor{tabsalmon!10}{54.15} & \cellcolor{tabsalmon!25}{46.22} & \cellcolor{tabsalmon!14}{53.45} & \cellcolor{tabsalmon!23}{19.65} \\
\addlinespace[3pt]
 & \textsc{ChatGPT}\textsubscript{Audio} & \cellcolor{tabsalmon!0}{10.04} & \cellcolor{tabsalmon!0}{17.42} & \cellcolor{tabsalmon!0}{14.61} & \cellcolor{tabsalmon!0}{17.24} & \cellcolor{tabsalmon!0}{15.36} & \cellcolor{tabsalmon!0}{5.66} & \cellcolor{tabsalmon!0}{17.24} & \cellcolor{tabsalmon!0}{3.41} \\
\midrule
\multicolumn{10}{c}{\textit{X $\to$ EN: Text-to-Text Translation}} \\
\midrule
\multirow{2}{*}{Cascaded} & \textsc{Llama}\textsubscript{Human transcript} & \cellcolor{tabblue!0}{52.89} & \cellcolor{tabblue!0}{62.62} & \cellcolor{tabblue!0}{63.21} & \cellcolor{tabblue!0}{66.02} & \cellcolor{tabblue!0}{65.16} & \cellcolor{tabblue!0}{49.75} & \cellcolor{tabblue!0}{66.29} & \cellcolor{tabblue!0}{53.90} \\
 & \textsc{Tower}\textsubscript{Human transcript} & \cellcolor{tabblue!4}{59.86} & \cellcolor{tabblue!50}{\textbf{68.97}} & \cellcolor{tabblue!17}{68.15} & \cellcolor{tabblue!2}{69.45} & \cellcolor{tabblue!3}{69.51} & \cellcolor{tabblue!21}{70.49} & \cellcolor{tabblue!3}{70.62} & \cellcolor{tabblue!26}{56.11} \\
\midrule
\multirow{2}{*}{Commercial} & \textsc{Gemini}\textsubscript{Human transcript} & \cellcolor{tabblue!50}{\textbf{67.53}} & \cellcolor{tabblue!36}{68.13} & \cellcolor{tabblue!50}{\textbf{69.14}} & \cellcolor{tabblue!50}{\textbf{78.26}} & \cellcolor{tabblue!50}{\textbf{77.04}} & \cellcolor{tabblue!50}{\textbf{73.46}} & \cellcolor{tabblue!50}{\textbf{76.51}} & \cellcolor{tabblue!50}{\textbf{57.36}} \\
 & \textsc{ChatGPT}\textsubscript{Human transcript} & \cellcolor{tabblue!37}{65.41} & \cellcolor{tabblue!6}{66.29} & \cellcolor{tabblue!46}{69.02} & \cellcolor{tabblue!39}{76.33} & \cellcolor{tabblue!34}{74.45} & \cellcolor{tabblue!23}{70.73} & \cellcolor{tabblue!31}{74.10} & \cellcolor{tabblue!2}{54.79} \\
\bottomrule
\end{tabular}%
}
\caption{\textbf{Cascaded systems lead among open-source models, and commercial systems achieve the highest scores overall.} chrF scores (standard prompt) for all models and transcript conditions, English$\to$X and X$\to$English. Subscripts: Audio = speech input; ASR = automatic transcript; Human = human transcript.}
\label{tab:chrf_main}
\end{table*}
We first examine general translation quality, before turning to disfluency preservation and an analysis of specific disfluency types.

\subsection{Comparison to Human Annotation and Automatic Metrics}\label{subsec:human_corr}

\Cref{tab:correlation} reports how well each automatic metric tracks human Style, Meaning, and Overall judgments (Overall being the average of the two) at both the segment level (Kendall's $\tau$) and the system level (soft pairwise accuracy). A per-language breakdown is shown \Cref{tab:correlation_per_language}. Unlike the LLM judge variants and the human protocol, chrF and COMET-Kiwi do not distinguish style from meaning: each produces a single quality score, which we correlate against the three human axes in turn. 

All four metrics correlate well with human judgment: system-level SPA exceeds 78\% across every metric and axis, and segment-level $\tau$ ranges from 27\% to 47\%, consistent with state-of-the-art results at recent speech translation metric shared tasks~\citep{adelani-etal-2026-speech}. The two reference-based metrics, chrF and LLM judge (ref), consistently outperform the reference-free metrics on every axis and at both granularities, in line with COMET-Kiwi being trained predominantly on fluent translation data~\citep{zufle2026needspeechevaluatespeech} and therefore less sensitive to disfluency removal when semantic content is preserved. Among the reference-based metrics, LLM judge (ref) attains the highest segment-level $\tau$ across all three axes, while at the system level the two are virtually indistinguishable. However, the LLM judge is only applied to the four systems selected for human evaluation and cannot rank all models in our benchmark. chrF captures character-level overlap with the disfluent reference, rewards outputs that retain disfluencies, and achieves the highest system-level SPA overall and for style. We therefore use it as our primary metric in the following.

\subsection{Translation Quality}
\label{sec:translation-quality}
 \Cref{tab:chrf_main} presents results on \textsc{Uh-Mazing} across architectures, using the standard prompt. Full results including the disfluency-aware prompt as well as the COMET  and LLM judge results are  given in Appendix~\ref{app:detailed_results}.

\paragraph{Architecture.}
We compare four architecture categories: cascaded, end-to-end, SpeechLLM, and commercial. The strongest systems overall are \textsc{Tower}\textsubscript{Human transcript} among open-source models and \textsc{ChatGPT}\textsubscript{ASR transcript} among commercial models, both using transcript input. Cascaded systems lead by a wide margin over all audio-input approaches, including end-to-end models, SpeechLLMs, and commercial audio models. 

Within cascaded systems, both ASR models have substantial WER against the disfluent reference (Whisper: $20.31\%$; Canary: $15.69\%$), yet their WER against the fluent reference is higher (Whisper: $22.27\%$; Canary: $24.99\%$), confirming that both preserve some disfluencies in transcription (details in \cref{tab:asr_results}. The gap is considerably larger for Canary ($9.3$pp vs. $2$pp for Whisper), indicating it retains more disfluency content, consistent with its lower E-Scores and Z-Scores. \textsc{Tower}\textsubscript{Canary} therefore outperforms \textsc{Tower}\textsubscript{Whisper} not only because of lower WER, but because Canary transcribes disfluencies more faithfully (details in \Cref{tab:asr_results}). Interestingly, ASR transcript input outperforms human transcript input for commercial models on all eight languages, whereas for open-source cascaded systems the expected pattern holds and human transcript yields the highest scores. \textsc{ChatGPT}\textsubscript{ASR transcript} outperforms all open-source systems, with one exception: \textsc{Tower}\textsubscript{Human transcript}, which uses a manually transcribed transcript rather than ASR output and is therefore not achievable in real deployment.

Among audio-input models, \textsc{Gemini}\textsubscript{Audio} outperforms end-to-end and SpeechLLM systems on most languages by a large margin.

\begin{figure*}
    \centering
    \includegraphics[width=1\linewidth]{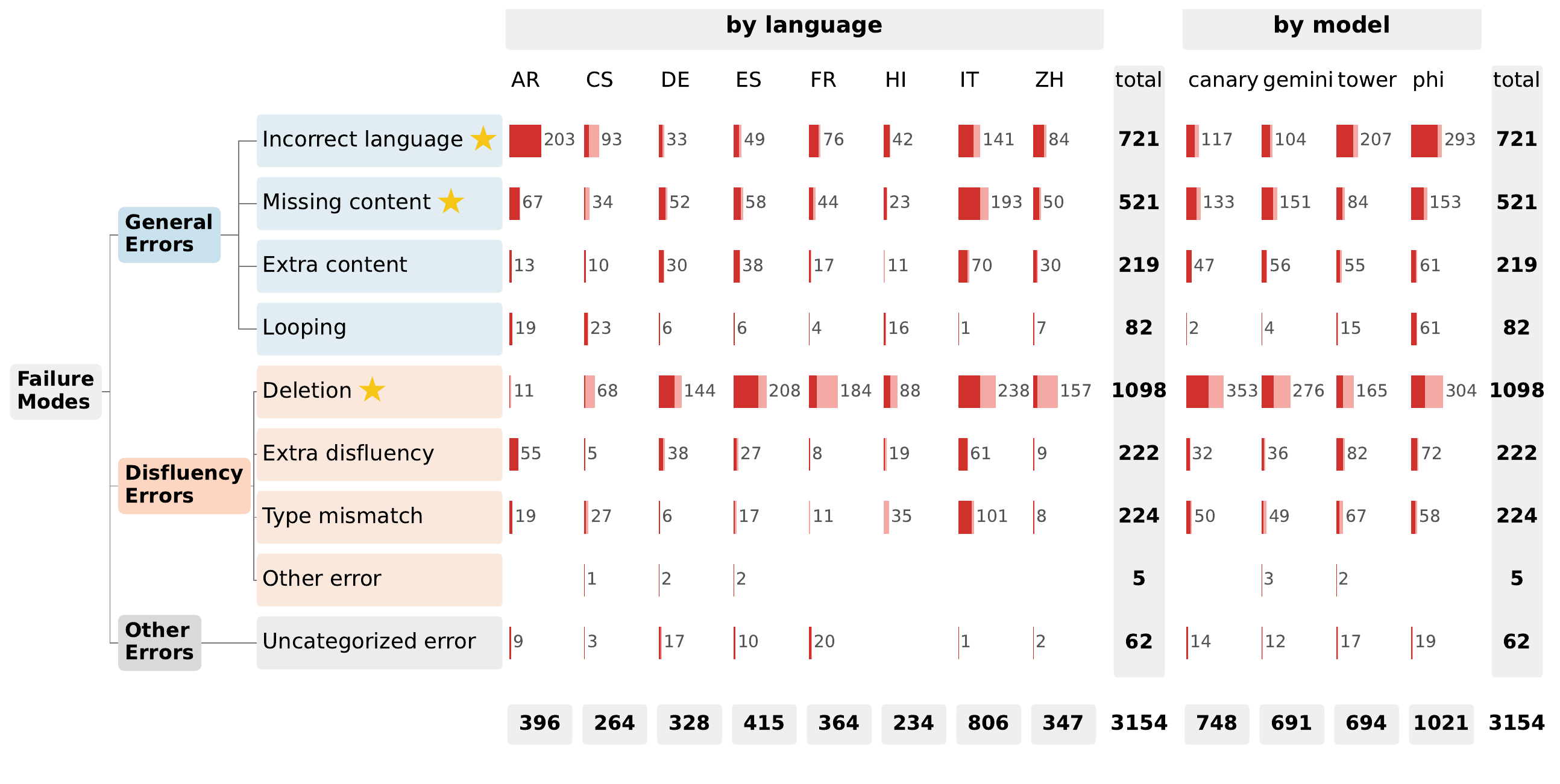}
    \caption{\textbf{Three failure modes  ({\color{yellow!80!orange}$\bigstar$}), 
    \textcolor[HTML]{2B81AB}{incorrect language}, 
    \textcolor[HTML]{2B81AB}{missing content}, and
    \textcolor[HTML]{D3591C}{deletion}, dominate consistently}, regardless of whether error spans are broken down by language (left) or model (right).
    Human annotators marked the spans, assigning each a category from our taxonomy and a severity (\textcolor{red!75!black}{\textbf{Major}}/\textcolor{red!40}{\textbf{Minor}}).}
    \label{fig:error-spans-and-taxonomy}
\end{figure*}

\paragraph{Prompt formulation.} The effect of explicitly asking models to preserve disfluencies is inconsistent across architectures (\Cref{tab:chrf} in the appendix). Within cascaded systems, \textsc{Tower} is largely unaffected, while \textsc{Llama} loses up to 17.3 chrF points on some languages. Commercial systems tend to improve instead, often substantially: \textsc{Gemini} gains under every condition and language, by as much as 38.6 points for \textsc{Gemini}\textsubscript{Audio} on Hindi, while \textsc{ChatGPT} improves under most conditions but stays roughly flat with an automatic transcript. SpeechLLMs diverge in opposite directions: \textsc{Phi-4} drops sharply on most languages, while \textsc{Qwen2.5} improves consistently.

\paragraph{X$\to$EN Translation.}
Our benchmark supports the reverse direction in the text setting, which we use to test whether the EN$\to$X gap reflects model weakness on lower-resource target languages rather than disfluency handling specifically. Using text input (no target-language audio exists for this direction), all three systems score substantially higher translating into English than out of it (bottom part of \Cref{tab:chrf_main}), consistent with English being a high-resource target language that models are trained to produce more reliably.

\subsection{Disfluency Preservation}\label{subsec:disfluency_preservation}
Having established overall translation quality (\Cref{sec:translation-quality}), we now turn to whether models preserve disfluencies, using the LLM judge and human evaluation protocols described in \Cref{sec:eval_methods}.

\begin{figure}
    \centering
    \includegraphics[width=1.0\linewidth]{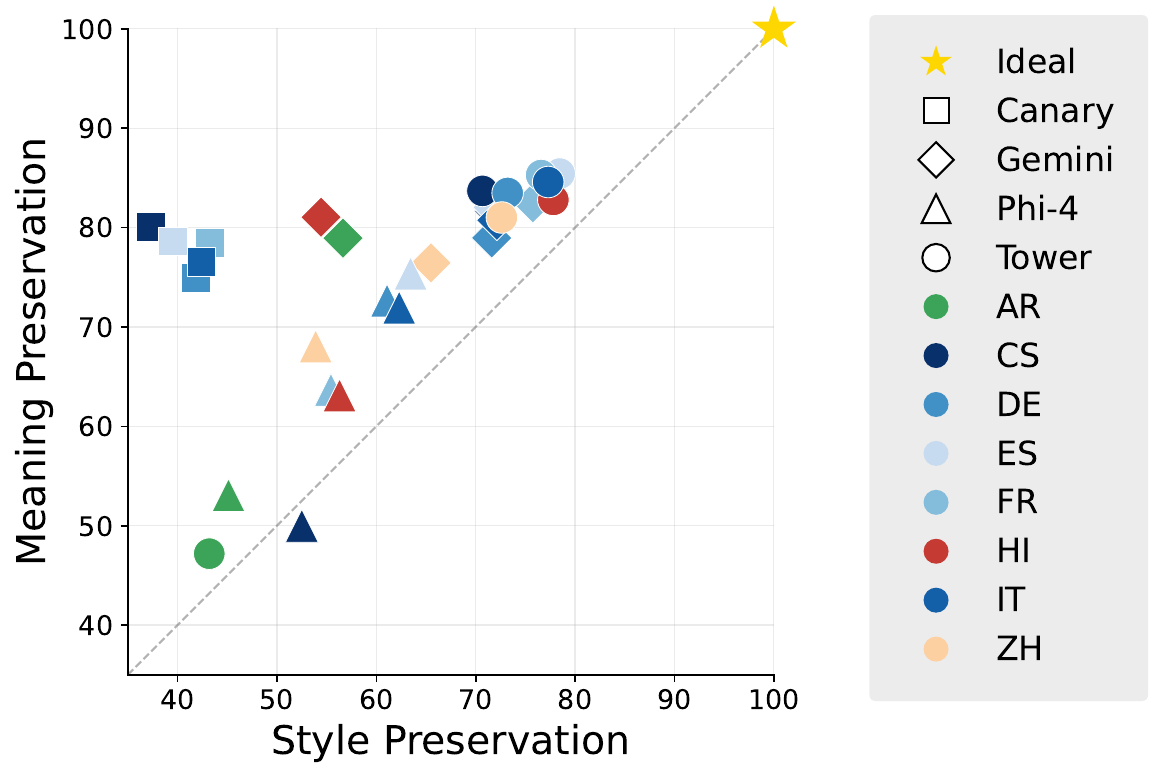}
    \caption{\textbf{All systems preserve meaning better than style in translation.} Style Preservation vs.\ Meaning Preservation scores for one representative system per architecture category, across eight languages.}
    \label{fig:style-vs-meaning}
\end{figure}

\paragraph{Style vs. Meaning.}

\Cref{fig:style-vs-meaning} plots the LLM judge Style and Meaning Preservation scores (0--100) for the best-performing system per architecture category, selected by overall translation quality (\Cref{sec:translation-quality}); full per-language scores are in \Cref{tab:llm_judge} in the appendix. Meaning preservation exceeds style preservation for nearly every system and language, with almost all points lying above the diagonal. The gap is largest for \textsc{Canary}, consistent with an end-to-end model that defaults to a fluent rendering rather than reproducing disfluencies, and smallest for \textsc{Tower}.

\subsection{Effect of Disfluency Type}\label{subsec:disfluency_type_ablation}
The previous section treated disfluencies as a single phenomenon, now we analyse different disfluency types and how they affect translation.

\paragraph{Failure Mode Taxonomy.}
To understand what kinds of errors occur, we built the failure-mode taxonomy in \Cref{fig:error-spans-and-taxonomy} from the error spans human annotators flagged during evaluation, together with a severity rating for each span. The taxonomy splits nearly evenly between its two top-level branches, General Errors (1,543 spans) and Disfluency Errors (1,549 spans), with a small remainder (62 spans) falling outside either. Three categories, incorrect language, missing content, and deletion of the source disfluency, account for nearly three-quarters of all 3,154 annotated spans, regardless of whether errors are broken down by language or by model. Deletion, where the model omits the disfluency rather than mistranslating it, is the single largest category on its own. Overall, models tend to simply ignore disfluencies rather than mistranslate them: deletions alone (1,098 spans) outnumber disfluency mistranslations, i.e.\ extra disfluencies plus type mismatches combined (446 spans). This pattern is broadly consistent across languages and models, with one notable exception: for Hindi, incorrect language, missing content, and deletion together account for only 65\% of spans, the lowest of any language (vs.\ 70--84\% elsewhere), driven by a comparably high share of disfluency type mismatches.

We repeat this breakdown for the four models with the highest chrF scores in their respective system categories: \textsc{Tower} (cascaded), \textsc{Canary} (end-to-end), \textsc{Phi-4} (SpeechLLM), and \textsc{Gemini} (commercial). The same three categories dominate for most models (73--81\% of spans); the exception is \textsc{Tower} (65.7\%), whose common errors are incorrect language rather than deletion.

\begin{figure}
    \centering
    \includegraphics[width=1.0\linewidth]{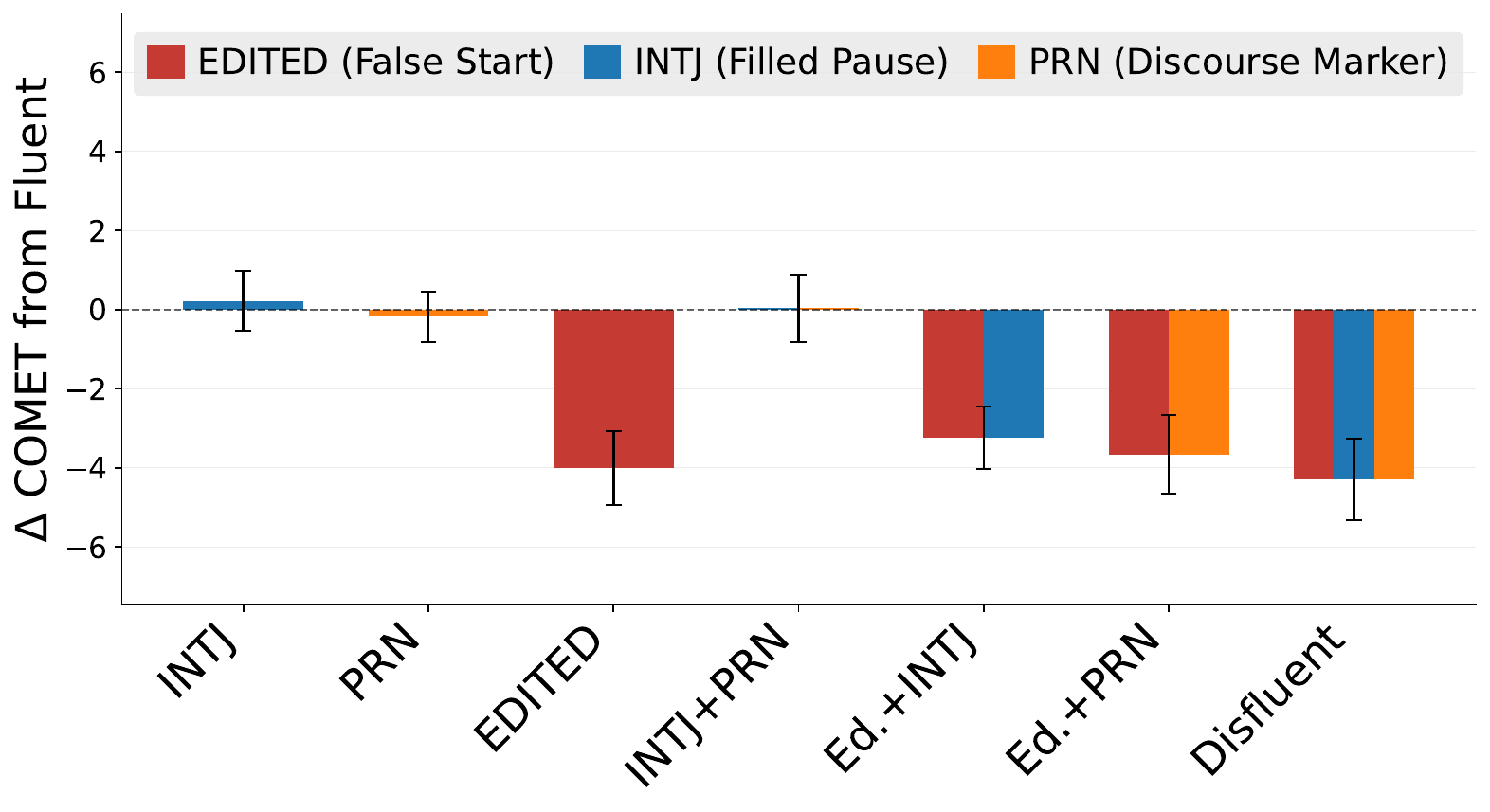}
    \vspace{-1mm}
    \caption{\textbf{EDITED-type disfluencies drive the drop in translation performance.} We show the change in COMET-Kiwi relative to the Fluent source baseline, by which disfluency  type(s) the source text contains, pooled across four translation models (Llama, Tower, Gemini, ChatGPT) and  eight target languages.}
    \label{fig:node_combos_delta}
\end{figure}

\paragraph{Quality by Disfluency Type.}
We distinguish three types annotated in Switchboard: INTJ (filled pauses), PRN (discourse markers), and EDITED (false starts and self-repairs). Since these types are text-level annotations, this ablation is restricted to text-input systems. We reinsert combinations of these types into the fluent human transcript and translate with all four text models (Llama, Tower, Gemini, and ChatGPT). \Cref{fig:node_combos_delta} shows the change in COMET-Kiwi relative to the fluent baseline: for each condition, we compute COMET-Kiwi on the disfluency-inserted source paired with its translation, and subtract the score for the fluent source paired with its translation. We find that filled pauses and discourse markers, alone or together, have no measurable effect on translation quality. Any condition containing a false start drops by roughly 3 to 4 points, regardless of what else is combined with it. The natural Disfluent transcript shows the same drop, confirming that false starts, not filler words or discourse markers, are responsible.

\section{Inference-Time Interventions}
\label{sec:Inference-Time Interventions}
Having established that current models fall short on disfluency preservation, we now ask whether performance can be improved without retraining since post-training for disfluency robustness risks reducing generalisation~\cite{teleki25_dres}, and annotated disfluent training data is scarce. We therefore investigate three lightweight inference-time strategies, beam search, temperature sampling, and in-context learning~\cite{dong24_survey, mei2025survey}, aimed at the failure modes from \cref{fig:error-spans-and-taxonomy}, without modifying model parameters.

\begin{figure}
    \centering
    \includegraphics[width=1.0\linewidth]{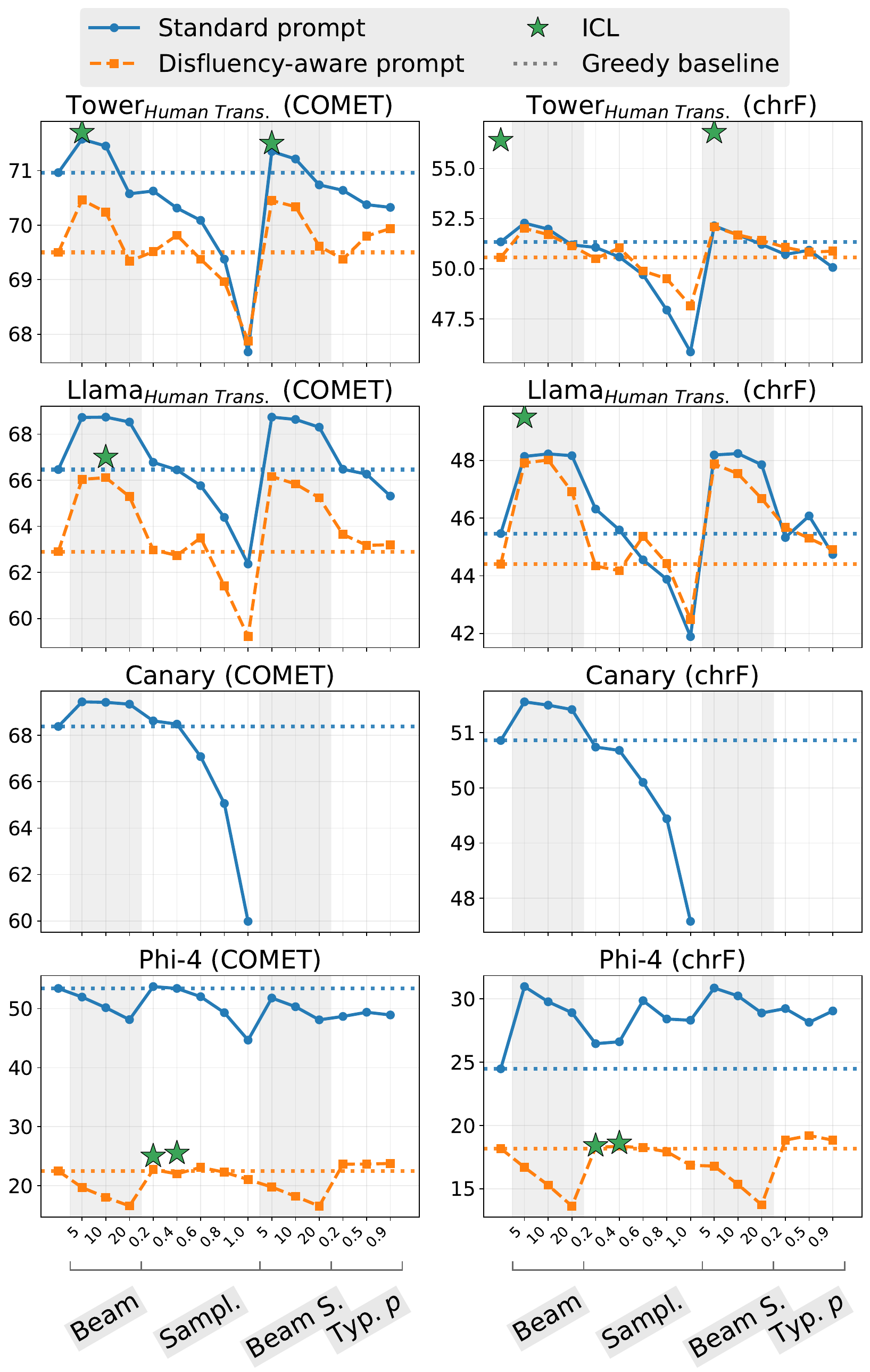}
\caption{\textbf{Decoding methods shape overall quality (COMET), while ICL improves disfluency preservation (chrF); the two interventions complement each other, as do the metrics.} Decoding strategies per model (rows), in \textbf{COMET} (left) and
\textbf{chrF} (right), averaged over eight EN$\to$X pairs. Lines compare the
\textcolor[HTML]{257BB6}{\textbf{standard}} (solid) and
\textcolor[HTML]{FF7F0E}{\textbf{disfluency-aware}} (dashed) prompts; dotted lines
are greedy baselines for the two prompt versions and {\color[HTML]{3BA458}$\bigstar$} marks
\textcolor[HTML]{3BA458}{\textbf{ICL}} runs.}
\label{fig:decoding-iclresults}
\end{figure}

\paragraph{Interventions.} We evaluate three families of intervention. First, we vary the decoding algorithm: greedy, beam search (beam size $n\!\in\!\{5,10,20\}$), temperature sampling ($\tau\!\in\!\{0.2,0.4,0.6,0.8,1.0\}$), beam sampling, and typical-$p$ sampling. Second, we compare the standard and disfluency-aware prompts of \cref{sec:eval} as an input-level intervention. Third, we add in-context learning \citep[ICL]{brown2020languagemodelsfewshotlearners}: for each test instance we sample $k\!=\!2$ demonstrations uniformly at random from the reference pool $\mathcal{D}$ ($N\!=\!80$) under a leave-one-out scheme that excludes the current item to prevent leakage, where each demonstration is a disfluent source paired with its disfluent reference translation. 

\paragraph{Inference-Time Interventions have impact on disfluent translation.}
\cref{fig:decoding-iclresults} reports the effect of these methods on overall quality in COMET and chrF, averaged over the eight EN$\to$X pairs. Across models, beam search consistently improves over greedy decoding, while quality falls as the sampling temperature rises. The two settings show complementary patterns: the disfluency-aware prompt leaves chrF largely unchanged but lowers COMET, whereas in-context learning barely moves COMET yet produces a meaningful improvement on chrF. For example, on Tower\textsubscript{Human Trans.}, ICL shifts COMET by less than 0.2 points while raising chrF by up to 5 points relative to the matched decoding setting, a pattern that recurs, in attenuated form, for Llama. Since chrF compares against the disfluent reference and is therefore sensitive to disfluency preservation (\Cref{subsec:human_corr}), this suggests that in-context learning specifically helps models preserve disfluencies without sacrificing meaning. \textsc{Phi-4} departs from these trends, though its overall quality is too low to draw reliable conclusions. Together, these results suggest that beam search and in-context learning are simple, low-risk ways to improve disfluency handling without sacrificing translation quality.

\section{Discussion and Conclusion}
\label{sec:discussion}
Taken together, our findings on the introduced \textsc{Uh-Mazing} benchmark make a connected argument: disfluencies carry emotional signal that current speech translation systems fail to preserve.

\paragraph{Disfluencies are meaningful, structured signals that are often not translated.}
The case study in \cref{sec:role_of_disfluencies} shows that removing disfluencies from speech shifts perceived negative emotion. Our analysis of \textsc{Uh-Mazing} shows disfluencies are not placed at random, systematically preceding the sentence subject and following a conjunction. Yet current systems, across four architecture types, preserve meaning reliably but fail on style (\cref{sec:Results}): our human-annotated error taxonomy shows this failure is dominated by deletion, with models omitting disfluencies rather than mistranslating them. This failure is concentrated in false starts and self-repairs specifically. Beam search and in-context learning narrow this style gap without retraining, but do not close it.

\paragraph{Implications for practitioners and researchers.}
The meaning-style gap identified in this work reflects the fact that current systems were trained to produce fluent translations, not to preserve the manner of delivery. Closing this gap will require training that treats disfluency preservation as a target in its own right. An open question is whether preserved disfluencies carry the same pragmatic and emotional meaning for target-language listeners as they do in the source. Studies at the intersection of translation and Human-Computer-Interaction could shed light on how translated hesitations and self-corrections are perceived across languages and cultures \cite{carpuat2025interdisciplinary, liebling2022opportunities}. We hope that our analysis and the release of \textsc{Uh-Mazing} provide a starting point for closing this gap.

\section{Generative AI Use Disclosure}
Generative AI tools were used to assist with drafting and refining portions of the manuscript, as well as for coding and figure preparation. All scientific content, analyses, and conclusions were developed and verified by the authors.

\section{Acknowledgments}
We thank Haoran Liu, Anwesha Basu, Majid Alfifi, Jobin Varughese, Xiangjue Dong, and Thomas Docog for expert annotation and valuable discussions. This work has received funding from the European Union’s Horizon research and innovation programme under grant agreement No 101135798, project Meetween (My Personal AI Mediator for Virtual MEETings BetWEEN People).

\bibliography{tacl2021}
\bibliographystyle{acl_natbib}


\onecolumn
\appendix
\appendix

\clearpage

\label{app:emotion}

\begin{figure}
\section{Human Case Study of the Meaning Preservation of Disfluencies}
    \centering
    \includegraphics[width=1.0\linewidth]{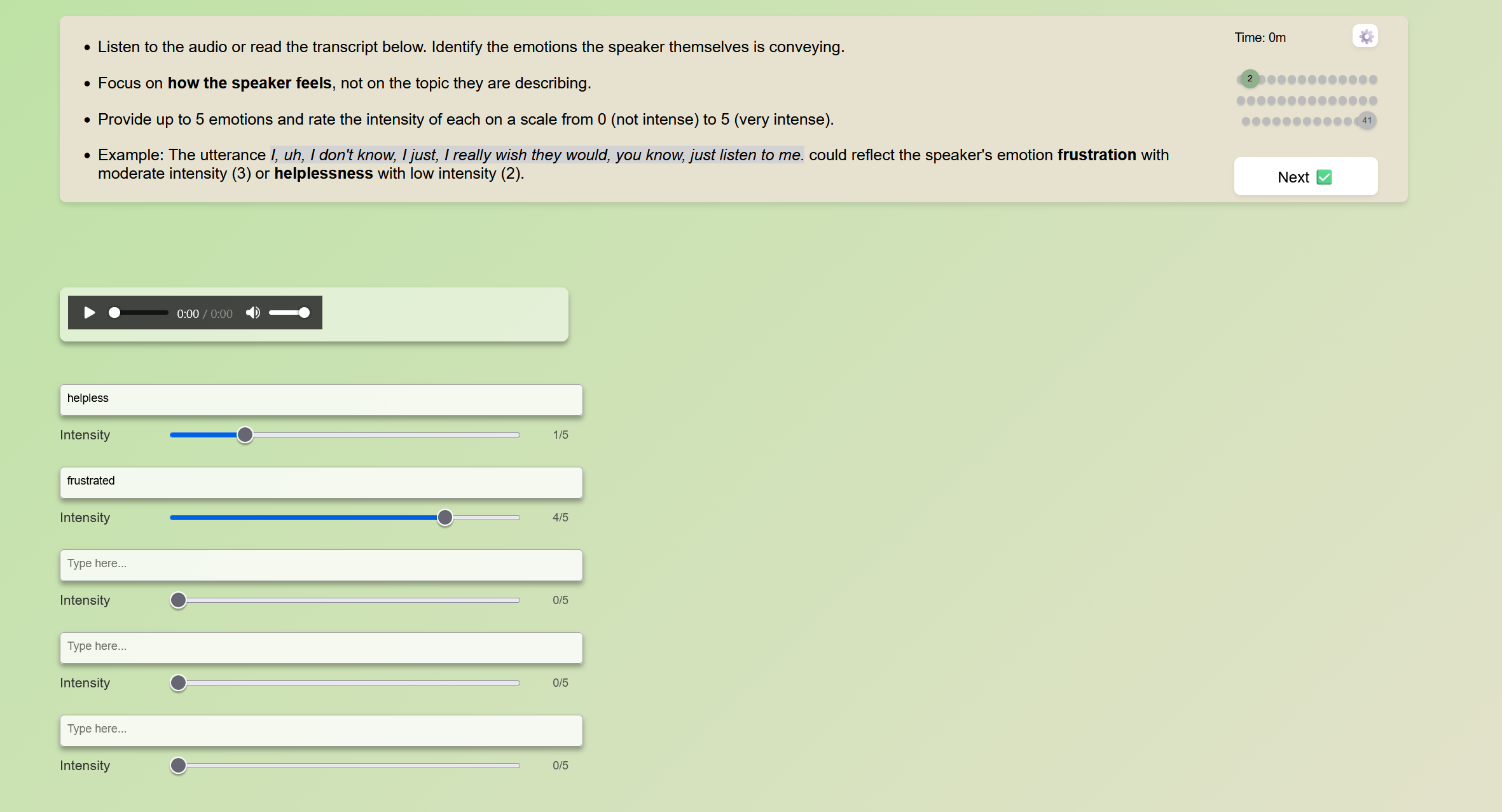}
    \caption{Instructions to the annotators to identify the meaning of disfluencies using the Pearmut platform \citep{zouhar2026pearmuthumanevaluationtranslation}. The same example was seen by different annotators using the source speech, the fluent transcript, and the disfluent transcript.}
    \label{fig:role_of_disfluency_pearmut}
\end{figure}

\begin{table*}[t]
\centering\small
\begin{tabular}{clp{5.8cm}p{5.8cm}}
\toprule
utt. & \makecell{neg-share\\A/D/F} & Disfluent transcript & Fluent transcript \\
\midrule
36 & \footnotesize 23/60/94\% & \emph{It would, it would be pirated and they wouldn't bother to check that carefully anyway to someone who's offering, you know, full cash price, for it.} & \emph{it would be pirated and they wouldn't bother to check that carefully anyway to someone who's offering full cash price, for it.} \\
\addlinespace
33 & \footnotesize 14/80/83\% & \emph{Um, the, the other side of that might be if, if someone found out something or surmised something that weren't true then I would feel probably more invaded in the gossipy sort of sense.} & \emph{the other side of that might be if someone found out something or surmised something that weren't true then I would feel probably more invaded in the gossipy sort of sense.} \\
\addlinespace
79 & \footnotesize 15/62/83\% & \emph{Yeah, they're, they're still around, they've got a ne-, new C D out, but I, I wouldn't buy it. Because see, what happens is, the old, see whe-, I like, I like the old Rolling Stones. I don't like the new stuff.} & \emph{they're still around, they've got a new C D out, but I wouldn't buy it. Because, what happens is, the old, whe-, I like the old Rolling Stones. I don't like the new stuff.} \\
\addlinespace
70 & \footnotesize 7/7/75\% & \emph{Well, I don't know, um, I, uh, have attended some seminars that had some tapes that went with them, but, uh, I guess not so much books although they sometimes have manuals and things, but, uh, they would be things on like how to be successful and sort of talking to yourself. You know ge-, getting your, yourself in gear to, uh, sort of pull yourself up by your boot straps and do what you really want to do. Convincing you that you need to get on with it.} & \emph{I don't know, I, have attended some seminars that had some tapes that went with them, but, not so much books although they sometimes have manuals and things, but, they would be things on how to be successful and sort of talking to yourself getting yourself in gear to, sort of pull yourself up by your boot straps and do what you really want to do. Convincing you that you need to get on with it.} \\
\addlinespace
37 & \footnotesize 33/73/100\% & \emph{I mean, you know, you can't tell it, what a company really has to do with it and there's something rather ominous about having virtually anyone, any hacker being able to know what your income is, what your spending habits are, and, you know, and, and that hacker just has to get in to, in touch with the sneak thief and suddenly and then what started as an invasion of privacy can be an invasion of your actual home.} & \emph{you can't tell it, what a company really has to do with it and there's something rather ominous about having virtually anyone, any hacker being able to know what your income is, what your spending habits are, and that hacker just has to get in touch with the sneak thief and suddenly what started as an invasion of privacy can be an invasion of your actual home.} \\
\addlinespace
\bottomrule
\end{tabular}
\caption{Utterances with the largest cross-condition swings in negative-label share (Audio\,/\,Disfluent\,/\,Fluent). Removing disfluencies replaces contemplative labels with hostile or distressed ones.}
\label{tab:qual-examples}
\end{table*}

\begin{figure*}
    \centering
    \includegraphics[width=1.0\linewidth]{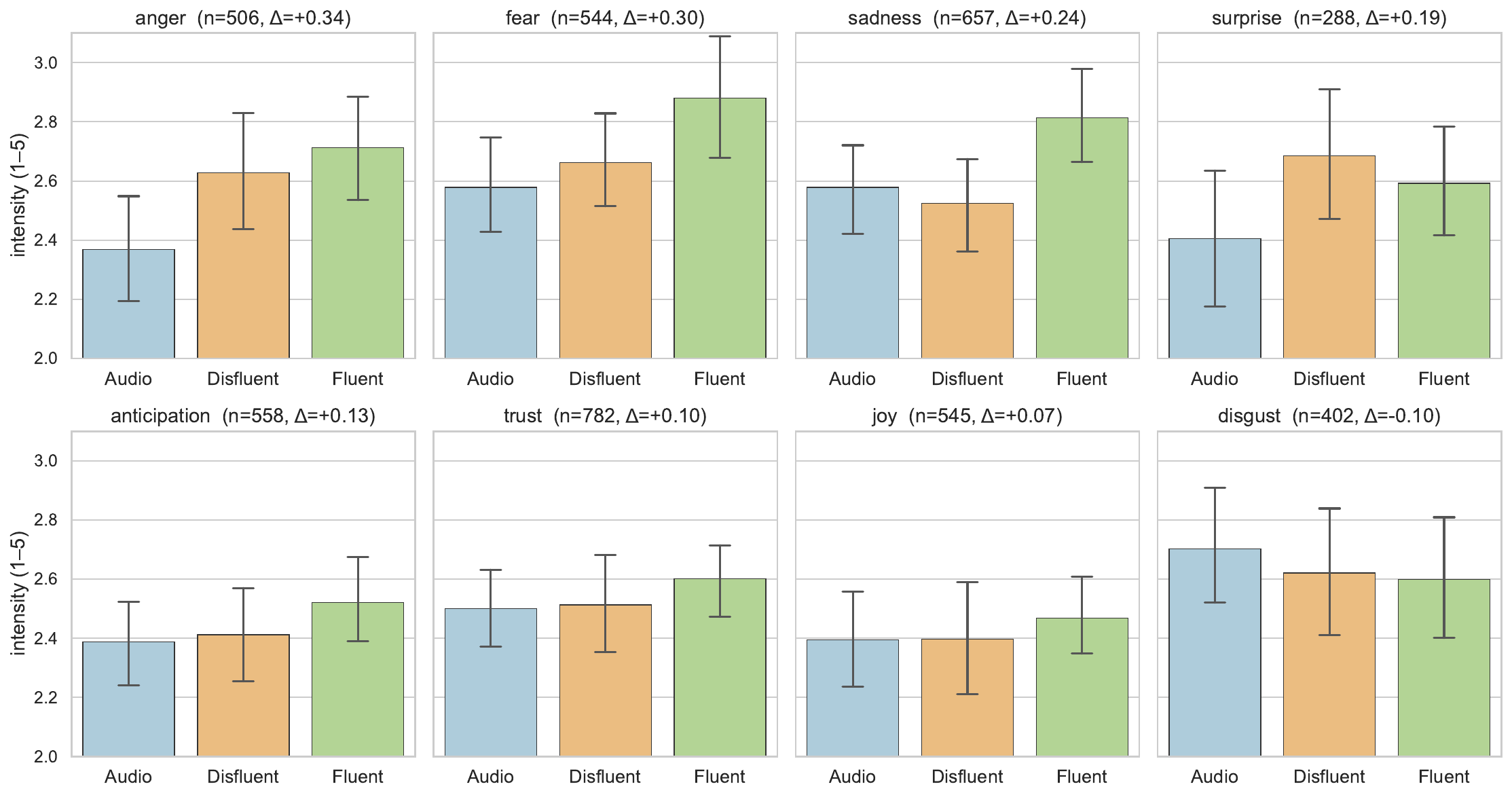}
    \caption{Mean intensity ratings per NRC emotion across the three conditions (Audio, Disfluent, Fluent), with 95\% confidence intervals. Each panel reports the total number of labels in that category ($n$) and the overall shift $\Delta_{\mathrm{F-A}} = \mathrm{Fluent} - \mathrm{Audio}$.}
    \label{fig:intensity_emotions}
\end{figure*}

\clearpage

\begin{figure}
\section{\textsc{Uh-Mazing} Benchmark Details}
\begin{minipage}{0.46\linewidth}
\begin{figure}[H]
  \centering
  \includegraphics[width=\linewidth]{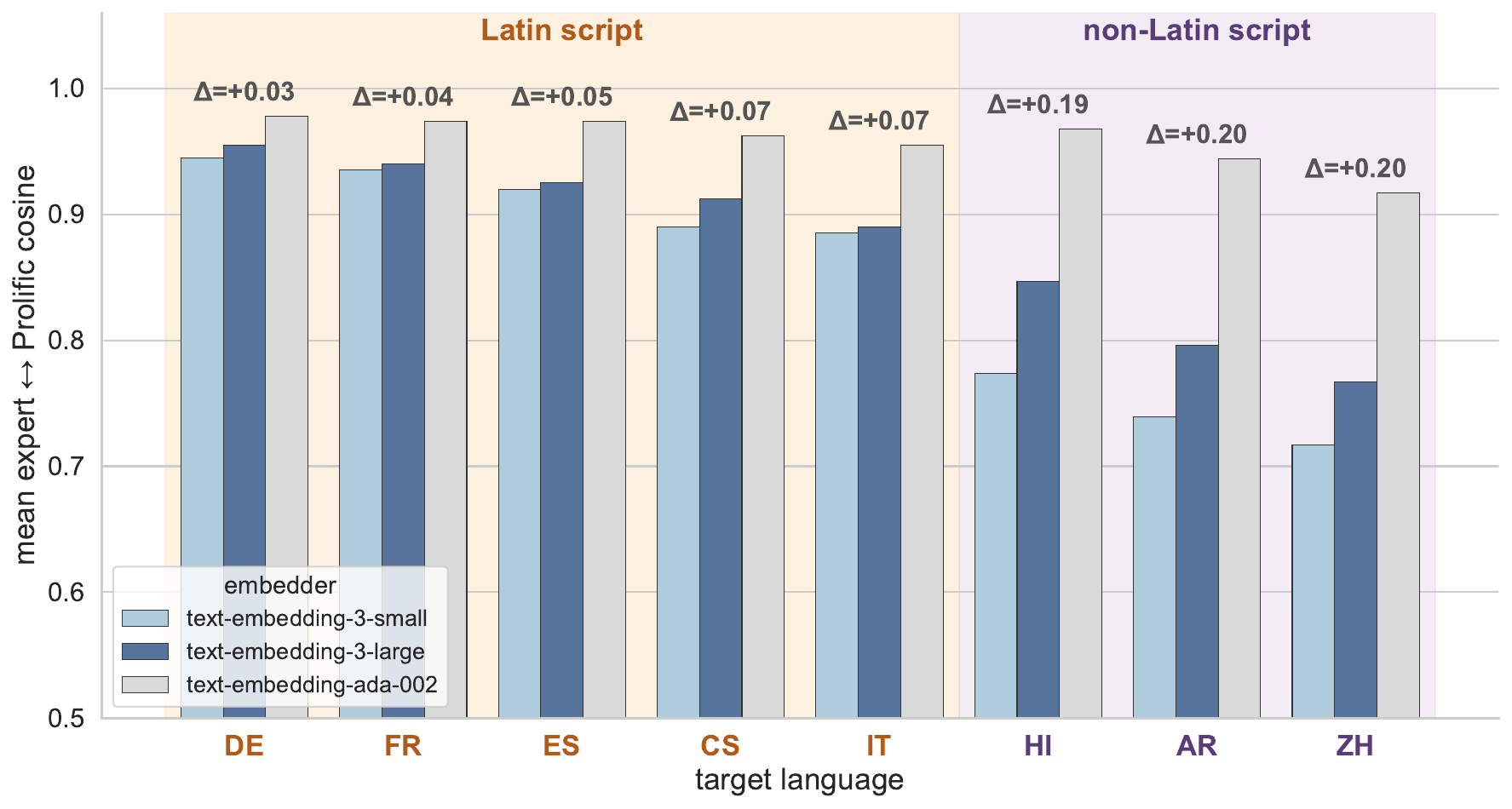}
  \caption{Mean expert$\leftrightarrow$Prolific cosine similarity per
    target language under three OpenAI embedders
    (\texttt{text-embedding-3-small}, \texttt{-3-large}, and
    \texttt{-ada-002}). The Latin- vs.\ non-Latin-script gap shrinks and
    eventually inverts as the embedder changes, indicating that part of
    the apparent non-Latin disagreement is embedder representational
    asymmetry rather than translator disagreement. $\Delta$ above each
    language shows the per-language change from
    \texttt{text-embedding-3-small} to \texttt{text-embedding-ada-002}.}
  \label{fig:translation-iaa}
\end{figure}
\end{minipage}
\hfill
\begin{minipage}{0.46\linewidth}
\begin{table}[H]
\centering
\small
\begin{tabular}{lrrr}
\toprule
\textbf{Lang} & \textbf{Utt.\ length} & \textbf{Disfl.\ length} & \textbf{\# Disfl.} \\
\midrule
AR & 56.3 & 2.7 & 452 \\
CS & 63.8 & 2.8 & 440 \\
DE & 70.3 & 3.6 & 428 \\
EN & 64.9 & 3.0 & 457 \\
ES & 68.6 & 3.1 & 452 \\
FR & 73.7 & 3.0 & 482 \\
HI & 76.2 & 3.2 & 459 \\
IT & 66.7 & 2.8 & 443 \\
ZH & 98.3 & 3.8 & 452 \\
\bottomrule
\end{tabular}
\caption{Average utterance length and disfluency length in words separated by whitespace, and total number of disfluencies (contiguous disfluency-marked spans: a single filler word and a 5-word false start each count as one), per language. Chinese lengths use jieba word-segmentations.}
\label{tab:basic-statistics}
\end{table}
\end{minipage}
\end{figure}

\begin{figure}
\begin{tcolorbox}[title=Round 1 Instructions: Translation, coltitle=black]
\textbf{Background:}
This study is conducted for research purposes. Data will be collected and analyzed as described in our Privacy Policy, linked here [link]. Participation is voluntary, and you may exit the study at any time without penalty. By continuing, you confirm that you have reviewed the Privacy Policy and consent to take part in this study.\\[1em]
\textbf{Task:}
In this task, you will act as a translator. You will be given short text passages written in a disfluent or non-standard form (e.g., grammatical errors, informal phrasing, incomplete sentences) and asked to translate them into the target language specified.

Translate only the text provided. Do not correct, rewrite, or add information beyond what is present in the original text unless explicitly instructed. Preserve the meaning as accurately as possible, even when the source text is unclear or awkward. The ID inside the square brackets (e.g., \textit{[sw02005\_A\_55]}) at the beginning of each translation can be ignored --- this is an internal identifier to be used by the researchers later.

Work independently on each text. Spend no more than 5 minutes per item.\\[1em]
\textbf{For each item:}
\begin{enumerate}[leftmargin=*, noitemsep, topsep=0.3em]
    \item Provide your best possible translation into the target language.
    \item If a portion of the text is unclear, ambiguous, or impossible to translate directly, translate it as faithfully as you can without guessing missing content.
    \item Use underscores (e.g., \textit{\_uh\_}) to mark disfluent tokens in the translated speech.
\end{enumerate}
\end{tcolorbox}

\vspace{1em}

\begin{tcolorbox}[title=Round 2 Instructions: Disfluency Highlighting, coltitle=black]
\textbf{Background:} 
These are transcripts of real spoken English conversations. Spoken language contains \textbf{disfluencies} --- the repetitions, fillers, and false starts that naturally occur when people speak. For example:
\vspace{0.5em}
\begin{mdframed}[backgroundcolor=gray!20,hidealllines=true,innerleftmargin=10pt,innerrightmargin=10pt,innertopmargin=8pt,innerbottommargin=8pt]
"\hlc{I mean}, \hlc{I} --- \hlc{I} think that, \hlc{you know}, we should..."
\end{mdframed}
\vspace{0.5em}
\noindent The highlighted words (\textit{I mean}, \textit{I}, \textit{you know}) are disfluent --- they add no meaning and are artefacts of speaking.\\[1em]
\textbf{What you'll see:} 
Each screen shows one utterance: the \textbf{English source} on top, with disfluent tokens already highlighted in \hlc{yellow}, and a \textbf{Spanish translation} below it.\\[1em]
\textbf{Your task}
\begin{itemize}[leftmargin=*, noitemsep, topsep=0.3em]
    \item Look at the \textbf{highlighted English words} --- these are the disfluencies.
    \item Find their equivalents in the \textbf{Spanish translation} and make sure they are highlighted too.
    \item \textbf{Click a plain word} to highlight it as disfluent.
    \item \textbf{Click a highlighted word} to remove the highlight if it shouldn't be there.
    \item If the translation text is missing words, mistranslated, or otherwise wrong, click \textbf{Edit text} to fix it --- then re-check the highlights.
    \item When the highlights in the translation match the English, click \textbf{Save \& Next}.
\end{itemize}
\vspace{1em}
\textbf{What counts as disfluent:} 
We have already identified the disfluencies in the English using a formal linguistic definition --- you don't need to judge whether something is disfluent. \textbf{Your only job is to find the equivalent words in the translation and make sure they are highlighted.} Whatever is highlighted in English should have a corresponding highlight in the translation.
\end{tcolorbox}
    \caption{Instructions given to annotators for the creation of \textsc{Uh-mazing}. In round 1, annotators translated disfluent English source text into Spanish, marking disfluent tokens with underscores. In round 2, annotators verified and corrected disfluency highlights in the translations.} 
    \label{fig:instructions}
\end{figure}

\clearpage

\begin{figure}[t]
\centering
\section{Speech Translation Model Details}
\begin{tcolorbox}
\small
\textbf{Standard:} ``Listen to the following audio and translate the speech into \textit{\{target\_language\}} text.''

\smallskip

\textbf{Disfluency-aware:} ``Listen to the following audio and translate the speech into \textit{\{target\_language\}} text, \hlc[pastelblue]{keeping any disfluencies (such as `um', `uh', repetitions, and hesitations) in the translation.}''

\smallskip

For text-input models, ``Listen to the following audio'' is replaced with ``Translate the transcribed speech.''
\end{tcolorbox}

\begin{tcolorbox}
\small
We have a speech audio and its translation. The speech audio contains disfluencies, such as ``uh,'' ``um,'' repeated words, self-corrections, etc.
Your task is to evaluate both:
\begin{itemize}[noitemsep,topsep=2pt]
    \item How well the disfluencies are preserved in the translation
    \item How natural these disfluencies sound in the target language
\end{itemize}
Specifically, you will rate the overall quality of each translation in two aspects on a 0--100 scale, using sliders:
\begin{itemize}[noitemsep,topsep=2pt]
    \item \textbf{Style Preservation:} How well are disfluencies preserved (presence, type, placement, spoken feel)?
    \item \textbf{Meaning Preservation:} How accurately and completely is the source meaning preserved?
\end{itemize}
Rate each from 0\% (poor) to 100\% (excellent).

\smallskip

\hlc[pastelblue]{Here is a reference for the translation: \texttt{\{reference\}}}

\smallskip

Return your evaluation in the following JSON format:
\begin{verbatim}
{
  "translation_{1-4}": {
    "style_preservation": <0-100>,
    "meaning_preservation": <0-100>
  }
}
\end{verbatim}

\smallskip

Translation {1-4}: \texttt{\{translation\_1\_text\}}
\end{tcolorbox}

\caption{\textbf{Top:} Translation prompts for cascaded models, SpeechLLMs, and commercial models. The \hlc[pastelblue]{highlighted clause} is added only in the disfluency-aware variant. \textbf{Bottom:} LLM-as-a-judge prompt for disfluency-aware translation evaluation. The \hlc[pastelblue]{highlighted line} is omitted in the reference-free version.\vspace{-1.5cm}}
\label{fig:prompts}
\end{figure}
\begin{figure}[h]
    \centering
    \vspace{-2mm}
    \includegraphics[width=1.0\linewidth]{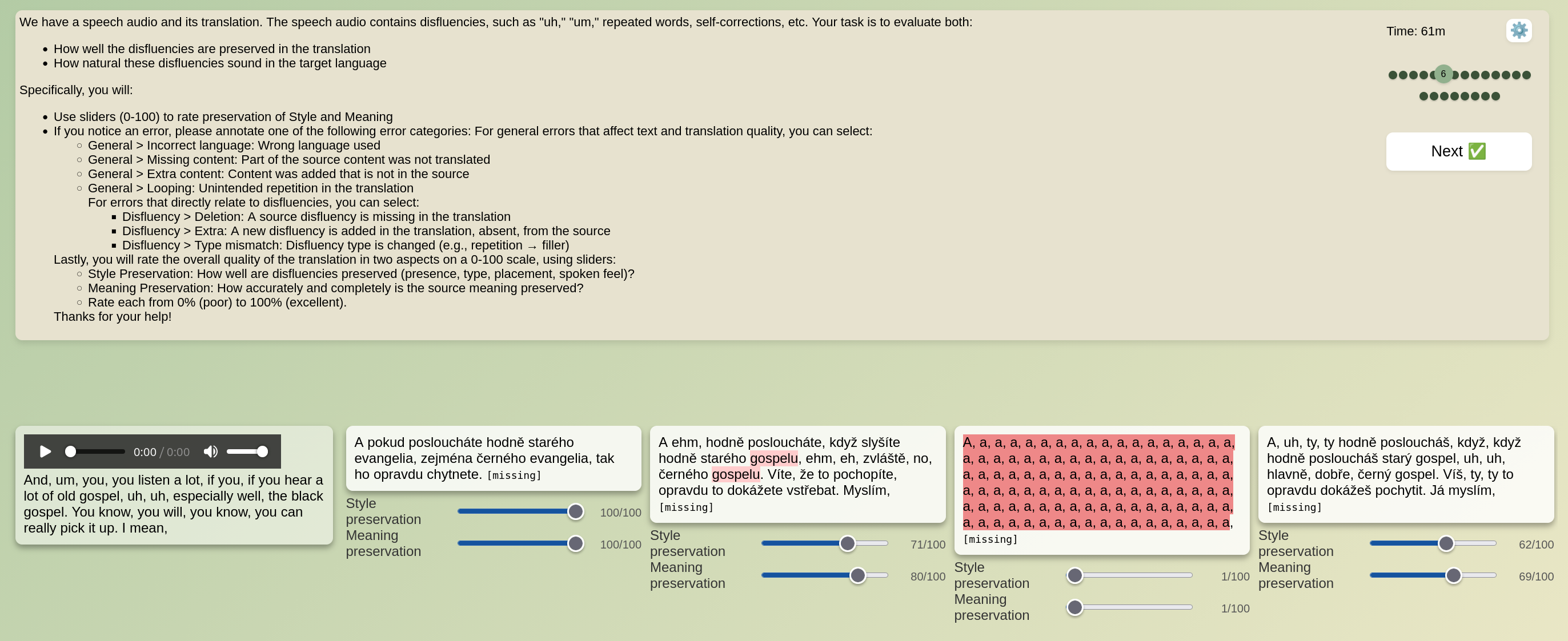}
    \vspace{-5mm}
    \caption{Annotation guidelines for marking and categorising error spans and general judgments of translation quality, in the Pearmut system \cite{zouhar2026pearmuthumanevaluationtranslation}.}
    \label{fig:pearmut_quality}
\end{figure}

\FloatBarrier

\begin{table*}[t]
\section{Results}\label{app:detailed_results}
\centering
\resizebox{\textwidth}{!}{%
\begin{tabular}{lllrrrrrrrr}
\toprule
\textbf{Category} & \textbf{Model} & \textbf{Prompt} & \textbf{AR} & \textbf{CS} & \textbf{DE} & \textbf{ES} & \textbf{FR} & \textbf{HI} & \textbf{IT} & \textbf{ZH} \\
\midrule
\multicolumn{11}{c}{\textit{EN $\to$ X Translations}} \\
\midrule
\multirow{12}{*}{Cascaded} & \multirow{2}{*}{\textsc{Llama}\textsubscript{Canary ASR}} & Std. & \cellcolor{tabsalmon!7}{22.16} & \cellcolor{tabsalmon!1}{38.37} & \cellcolor{tabsalmon!3}{49.30} & \cellcolor{tabsalmon!5}{52.53} & \cellcolor{tabsalmon!0}{52.12} & \cellcolor{tabsalmon!1}{35.91} & \cellcolor{tabsalmon!2}{50.06} & \cellcolor{tabsalmon!0}{13.42} \\
 &  & Disfl. & \cellcolor{tabsalmon!1}{18.96} & \cellcolor{tabsalmon!0}{31.67} & \cellcolor{tabsalmon!0}{42.07} & \cellcolor{tabsalmon!7}{53.20} & \cellcolor{tabsalmon!2}{53.12} & \cellcolor{tabsalmon!1}{35.98} & \cellcolor{tabsalmon!2}{50.21} & \cellcolor{tabsalmon!0}{13.10} \\
 & \multirow{2}{*}{\textsc{Llama}\textsubscript{Whisper ASR}} & Std. & \cellcolor{tabsalmon!22}{30.02} & \cellcolor{tabsalmon!0}{37.96} & \cellcolor{tabsalmon!0}{48.19} & \cellcolor{tabsalmon!3}{51.78} & \cellcolor{tabsalmon!0}{51.69} & \cellcolor{tabsalmon!7}{38.99} & \cellcolor{tabsalmon!0}{49.11} & \cellcolor{tabsalmon!0}{13.59} \\
 &  & Disfl. & \cellcolor{tabsalmon!0}{18.36} & \cellcolor{tabsalmon!5}{39.79} & \cellcolor{tabsalmon!4}{49.33} & \cellcolor{tabsalmon!0}{46.99} & \cellcolor{tabsalmon!0}{52.54} & \cellcolor{tabsalmon!0}{35.15} & \cellcolor{tabsalmon!0}{45.04} & \cellcolor{tabsalmon!0}{12.84} \\
\addlinespace[3pt]
 & \multirow{2}{*}{\textsc{Llama}\textsubscript{Human transcript}} & Std. & \cellcolor{tabsalmon!0}{16.64} & \cellcolor{tabsalmon!15}{43.97} & \cellcolor{tabsalmon!23}{56.30} & \cellcolor{tabsalmon!34}{61.40} & \cellcolor{tabsalmon!20}{59.29} & \cellcolor{tabsalmon!0}{33.25} & \cellcolor{tabsalmon!26}{59.11} & \cellcolor{tabsalmon!10}{17.07} \\
 &  & Disfl. & \cellcolor{tabsalmon!0}{14.49} & \cellcolor{tabsalmon!0}{37.60} & \cellcolor{tabsalmon!0}{39.00} & \cellcolor{tabsalmon!0}{49.21} & \cellcolor{tabsalmon!13}{57.13} & \cellcolor{tabsalmon!6}{38.43} & \cellcolor{tabsalmon!0}{47.78} & \cellcolor{tabsalmon!3}{14.72} \\
\addlinespace[3pt]
 & \multirow{2}{*}{\textsc{Tower}\textsubscript{Canary ASR}} & Std. & \cellcolor{tabsalmon!8}{22.87} & \cellcolor{tabsalmon!16}{44.43} & \cellcolor{tabsalmon!16}{53.86} & \cellcolor{tabsalmon!11}{54.40} & \cellcolor{tabsalmon!9}{55.67} & \cellcolor{tabsalmon!32}{52.24} & \cellcolor{tabsalmon!10}{53.21} & \cellcolor{tabsalmon!22}{21.13} \\
 &  & Disfl. & \cellcolor{tabsalmon!6}{21.66} & \cellcolor{tabsalmon!19}{45.91} & \cellcolor{tabsalmon!17}{54.33} & \cellcolor{tabsalmon!14}{55.14} & \cellcolor{tabsalmon!13}{56.97} & \cellcolor{tabsalmon!28}{50.15} & \cellcolor{tabsalmon!15}{55.20} & \cellcolor{tabsalmon!19}{20.26} \\
 & \multirow{2}{*}{\textsc{Tower}\textsubscript{Whisper ASR}} & Std. & \cellcolor{tabsalmon!9}{23.42} & \cellcolor{tabsalmon!14}{43.90} & \cellcolor{tabsalmon!10}{51.76} & \cellcolor{tabsalmon!8}{53.40} & \cellcolor{tabsalmon!2}{53.03} & \cellcolor{tabsalmon!32}{51.93} & \cellcolor{tabsalmon!6}{51.87} & \cellcolor{tabsalmon!20}{20.68} \\
 &  & Disfl. & \cellcolor{tabsalmon!3}{20.36} & \cellcolor{tabsalmon!15}{44.20} & \cellcolor{tabsalmon!12}{52.48} & \cellcolor{tabsalmon!9}{53.79} & \cellcolor{tabsalmon!8}{55.09} & \cellcolor{tabsalmon!30}{51.00} & \cellcolor{tabsalmon!8}{52.57} & \cellcolor{tabsalmon!17}{19.70} \\
\addlinespace[3pt]
 & \multirow{2}{*}{\textsc{Tower}\textsubscript{Human transcript}} & Std. & \cellcolor{tabsalmon!7}{22.39} & \cellcolor{tabsalmon!36}{52.90} & \cellcolor{tabsalmon!37}{61.34} & \cellcolor{tabsalmon!43}{64.05} & \cellcolor{tabsalmon!32}{63.65} & \cellcolor{tabsalmon!50}{61.43} & \cellcolor{tabsalmon!36}{62.93} & \cellcolor{tabsalmon!36}{25.89} \\
 &  & Disfl. & \cellcolor{tabsalmon!1}{19.37} & \cellcolor{tabsalmon!38}{53.88} & \cellcolor{tabsalmon!40}{62.23} & \cellcolor{tabsalmon!45}{64.67} & \cellcolor{tabsalmon!39}{66.17} & \cellcolor{tabsalmon!47}{59.82} & \cellcolor{tabsalmon!24}{58.42} & \cellcolor{tabsalmon!36}{25.92} \\
\midrule
\multirow{2}{*}{End-to-end} & \textsc{OWSM} & N/A & \cellcolor{tabsalmon!15}{26.47} & \cellcolor{tabsalmon!0}{27.38} & \cellcolor{tabsalmon!0}{47.76} & \cellcolor{tabsalmon!0}{43.02} & \cellcolor{tabsalmon!0}{43.17} & -- & \cellcolor{tabsalmon!0}{40.93} & \cellcolor{tabsalmon!4}{15.37} \\
 & \textsc{Canary} & N/A & -- & \cellcolor{tabsalmon!14}{43.89} & \cellcolor{tabsalmon!11}{52.13} & \cellcolor{tabsalmon!2}{51.65} & \cellcolor{tabsalmon!5}{54.30} & -- & \cellcolor{tabsalmon!9}{52.80} & -- \\
\midrule
\multirow{4}{*}{SpeechLLM} & \multirow{2}{*}{\textsc{Phi-4-MM-instruct}} & Std. & \cellcolor{tabsalmon!0}{5.73} & \cellcolor{tabsalmon!0}{8.01} & \cellcolor{tabsalmon!0}{45.79} & \cellcolor{tabsalmon!0}{46.03} & \cellcolor{tabsalmon!0}{27.12} & \cellcolor{tabsalmon!0}{3.56} & \cellcolor{tabsalmon!0}{49.58} & \cellcolor{tabsalmon!1}{14.28} \\
 &  & Disfl. & \cellcolor{tabsalmon!0}{0.71} & \cellcolor{tabsalmon!0}{13.52} & \cellcolor{tabsalmon!0}{20.84} & \cellcolor{tabsalmon!0}{27.21} & \cellcolor{tabsalmon!0}{20.44} & \cellcolor{tabsalmon!0}{1.42} & \cellcolor{tabsalmon!0}{37.79} & \cellcolor{tabsalmon!0}{8.97} \\
 & \multirow{2}{*}{\textsc{Qwen2.5-Omni}} & Std. & \cellcolor{tabsalmon!0}{17.32} & \cellcolor{tabsalmon!0}{10.61} & \cellcolor{tabsalmon!0}{20.90} & \cellcolor{tabsalmon!0}{18.02} & \cellcolor{tabsalmon!0}{18.27} & \cellcolor{tabsalmon!0}{11.78} & \cellcolor{tabsalmon!0}{21.28} & \cellcolor{tabsalmon!0}{12.65} \\
 &  & Disfl. & \cellcolor{tabsalmon!0}{17.68} & \cellcolor{tabsalmon!0}{15.06} & \cellcolor{tabsalmon!0}{33.04} & \cellcolor{tabsalmon!0}{28.29} & \cellcolor{tabsalmon!0}{24.48} & \cellcolor{tabsalmon!0}{12.38} & \cellcolor{tabsalmon!0}{27.11} & \cellcolor{tabsalmon!6}{15.74} \\
\midrule
\multirow{12}{*}{Commercial} & \multirow{2}{*}{\textsc{Gemini}\textsubscript{Human transcript}} & Std. & \cellcolor{tabsalmon!37}{38.00} & \cellcolor{tabsalmon!20}{46.18} & \cellcolor{tabsalmon!13}{52.62} & \cellcolor{tabsalmon!5}{52.51} & \cellcolor{tabsalmon!3}{53.44} & \cellcolor{tabsalmon!23}{47.18} & \cellcolor{tabsalmon!11}{53.60} & \cellcolor{tabsalmon!18}{19.87} \\
 &  & Disfl. & \cellcolor{tabsalmon!40}{39.52} & \cellcolor{tabsalmon!30}{50.39} & \cellcolor{tabsalmon!25}{57.10} & \cellcolor{tabsalmon!22}{57.75} & \cellcolor{tabsalmon!21}{59.94} & \cellcolor{tabsalmon!29}{50.55} & \cellcolor{tabsalmon!28}{60.02} & \cellcolor{tabsalmon!27}{22.80} \\
 & \multirow{2}{*}{\textsc{ChatGPT}\textsubscript{Human transcript}} & Std. & \cellcolor{tabsalmon!44}{41.74} & \cellcolor{tabsalmon!19}{45.82} & \cellcolor{tabsalmon!16}{53.97} & \cellcolor{tabsalmon!8}{53.35} & \cellcolor{tabsalmon!5}{54.15} & \cellcolor{tabsalmon!21}{46.22} & \cellcolor{tabsalmon!11}{53.45} & \cellcolor{tabsalmon!17}{19.65} \\
 &  & Disfl. & \cellcolor{tabsalmon!35}{36.93} & \cellcolor{tabsalmon!34}{51.94} & \cellcolor{tabsalmon!28}{58.08} & \cellcolor{tabsalmon!26}{58.96} & \cellcolor{tabsalmon!23}{60.63} & \cellcolor{tabsalmon!28}{49.72} & \cellcolor{tabsalmon!25}{58.71} & \cellcolor{tabsalmon!26}{22.73} \\
\addlinespace[3pt]
 & \multirow{2}{*}{\textsc{Gemini}\textsubscript{ASR transcript}} & Std. & \cellcolor{tabsalmon!42}{40.51} & \cellcolor{tabsalmon!37}{53.27} & \cellcolor{tabsalmon!36}{60.91} & \cellcolor{tabsalmon!31}{60.46} & \cellcolor{tabsalmon!30}{62.95} & \cellcolor{tabsalmon!36}{54.30} & \cellcolor{tabsalmon!33}{61.91} & \cellcolor{tabsalmon!30}{23.79} \\
 &  & Disfl. & \cellcolor{tabsalmon!45}{42.25} & \cellcolor{tabsalmon!50}{\textbf{58.74}} & \cellcolor{tabsalmon!50}{\textbf{65.99}} & \cellcolor{tabsalmon!50}{\textbf{66.23}} & \cellcolor{tabsalmon!50}{\textbf{69.96}} & \cellcolor{tabsalmon!50}{\textbf{61.48}} & \cellcolor{tabsalmon!50}{\textbf{68.31}} & \cellcolor{tabsalmon!50}{\textbf{30.67}} \\
 & \multirow{2}{*}{\textsc{ChatGPT}\textsubscript{ASR transcript}} & Std. & \cellcolor{tabsalmon!50}{\textbf{44.97}} & \cellcolor{tabsalmon!39}{54.19} & \cellcolor{tabsalmon!40}{62.28} & \cellcolor{tabsalmon!33}{60.92} & \cellcolor{tabsalmon!28}{62.12} & \cellcolor{tabsalmon!40}{56.17} & \cellcolor{tabsalmon!29}{60.34} & \cellcolor{tabsalmon!37}{26.42} \\
 &  & Disfl. & \cellcolor{tabsalmon!43}{41.12} & \cellcolor{tabsalmon!44}{56.14} & \cellcolor{tabsalmon!39}{61.95} & \cellcolor{tabsalmon!36}{61.98} & \cellcolor{tabsalmon!31}{63.27} & \cellcolor{tabsalmon!38}{55.07} & \cellcolor{tabsalmon!33}{61.97} & \cellcolor{tabsalmon!47}{29.73} \\
\addlinespace[3pt]
 & \multirow{2}{*}{\textsc{Gemini}\textsubscript{Audio}} & Std. & \cellcolor{tabsalmon!35}{37.15} & \cellcolor{tabsalmon!27}{48.97} & \cellcolor{tabsalmon!20}{55.24} & \cellcolor{tabsalmon!25}{58.54} & \cellcolor{tabsalmon!13}{56.83} & \cellcolor{tabsalmon!0}{15.90} & \cellcolor{tabsalmon!22}{57.60} & \cellcolor{tabsalmon!0}{7.13} \\
 &  & Disfl. & \cellcolor{tabsalmon!44}{41.62} & \cellcolor{tabsalmon!35}{52.66} & \cellcolor{tabsalmon!39}{62.11} & \cellcolor{tabsalmon!37}{62.29} & \cellcolor{tabsalmon!30}{62.79} & \cellcolor{tabsalmon!37}{54.51} & \cellcolor{tabsalmon!36}{62.86} & \cellcolor{tabsalmon!33}{25.05} \\
 & \multirow{2}{*}{\textsc{ChatGPT}\textsubscript{Audio}} & Std. & \cellcolor{tabsalmon!0}{10.04} & \cellcolor{tabsalmon!0}{17.42} & \cellcolor{tabsalmon!0}{14.61} & \cellcolor{tabsalmon!0}{17.24} & \cellcolor{tabsalmon!0}{15.36} & \cellcolor{tabsalmon!0}{5.66} & \cellcolor{tabsalmon!0}{17.24} & \cellcolor{tabsalmon!0}{3.41} \\
 &  & Disfl. & \cellcolor{tabsalmon!0}{8.13} & \cellcolor{tabsalmon!0}{19.06} & \cellcolor{tabsalmon!0}{18.83} & \cellcolor{tabsalmon!0}{24.32} & \cellcolor{tabsalmon!0}{20.84} & \cellcolor{tabsalmon!0}{15.15} & \cellcolor{tabsalmon!0}{19.48} & \cellcolor{tabsalmon!0}{5.63} \\
\midrule
\multicolumn{11}{c}{\textit{X $\to$ EN Translations}} \\
\midrule
\multirow{2}{*}{Cascaded} & \textsc{Llama}\textsubscript{Human transcript} & Std. & \cellcolor{tabblue!0}{52.89} & \cellcolor{tabblue!0}{62.62} & \cellcolor{tabblue!0}{63.21} & \cellcolor{tabblue!0}{66.02} & \cellcolor{tabblue!0}{65.16} & \cellcolor{tabblue!0}{49.75} & \cellcolor{tabblue!0}{66.29} & \cellcolor{tabblue!0}{53.90} \\
 & \textsc{Tower}\textsubscript{Human transcript} & Std. & \cellcolor{tabblue!4}{59.86} & \cellcolor{tabblue!50}{\textbf{68.97}} & \cellcolor{tabblue!17}{68.15} & \cellcolor{tabblue!2}{69.45} & \cellcolor{tabblue!3}{69.51} & \cellcolor{tabblue!21}{70.49} & \cellcolor{tabblue!3}{70.62} & \cellcolor{tabblue!26}{56.11} \\
\midrule
\multirow{2}{*}{Commercial} & \textsc{ChatGPT}\textsubscript{Human transcript} & Std. & \cellcolor{tabblue!37}{65.41} & \cellcolor{tabblue!6}{66.29} & \cellcolor{tabblue!46}{69.02} & \cellcolor{tabblue!39}{76.33} & \cellcolor{tabblue!34}{74.45} & \cellcolor{tabblue!23}{70.73} & \cellcolor{tabblue!31}{74.10} & \cellcolor{tabblue!2}{54.79} \\
 & \textsc{Gemini}\textsubscript{Human transcript} & Std. & \cellcolor{tabblue!50}{\textbf{67.53}} & \cellcolor{tabblue!36}{68.13} & \cellcolor{tabblue!50}{\textbf{69.14}} & \cellcolor{tabblue!50}{\textbf{78.26}} & \cellcolor{tabblue!50}{\textbf{77.04}} & \cellcolor{tabblue!50}{\textbf{73.46}} & \cellcolor{tabblue!50}{\textbf{76.51}} & \cellcolor{tabblue!50}{\textbf{57.36}} \\
\bottomrule
\end{tabular}%
}
\caption{chrF scores across models, prompts, and target languages for English$\to$X and X$\to$English translation. Std.\ = standard prompt; Disfl.\ = disfluency-aware prompt.}
\label{tab:chrf}
\end{table*}
\begin{table*}[t]
\centering
\resizebox{\textwidth}{!}{%
\begin{tabular}{lllrrrrrrrr}
\toprule
\textbf{Category} & \textbf{Model} & \textbf{Prompt} & \textbf{AR} & \textbf{CS} & \textbf{DE} & \textbf{ES} & \textbf{FR} & \textbf{HI} & \textbf{IT} & \textbf{ZH} \\
\midrule
\multicolumn{11}{c}{\textit{EN $\to$ X Translations}} \\
\midrule
\multirow{12}{*}{Cascaded} & \multirow{2}{*}{\textsc{Llama}\textsubscript{Canary ASR}} & Std. & \cellcolor{tabsalmon!17}{58.11} & \cellcolor{tabsalmon!5}{64.96} & \cellcolor{tabsalmon!6}{66.78} & \cellcolor{tabsalmon!10}{70.30} & \cellcolor{tabsalmon!11}{72.02} & \cellcolor{tabsalmon!1}{67.40} & \cellcolor{tabsalmon!11}{70.83} & \cellcolor{tabsalmon!8}{66.22} \\
 &  & Disfl. & \cellcolor{tabsalmon!15}{57.29} & \cellcolor{tabsalmon!0}{63.55} & \cellcolor{tabsalmon!0}{65.57} & \cellcolor{tabsalmon!0}{68.11} & \cellcolor{tabsalmon!2}{70.38} & \cellcolor{tabsalmon!0}{65.45} & \cellcolor{tabsalmon!0}{68.19} & \cellcolor{tabsalmon!0}{64.04} \\
 & \multirow{2}{*}{\textsc{Llama}\textsubscript{Whisper ASR}} & Std. & \cellcolor{tabsalmon!20}{59.46} & \cellcolor{tabsalmon!2}{64.22} & \cellcolor{tabsalmon!0}{65.40} & \cellcolor{tabsalmon!9}{70.16} & \cellcolor{tabsalmon!7}{71.33} & \cellcolor{tabsalmon!0}{67.18} & \cellcolor{tabsalmon!12}{71.04} & \cellcolor{tabsalmon!6}{65.92} \\
 &  & Disfl. & \cellcolor{tabsalmon!12}{55.64} & \cellcolor{tabsalmon!0}{63.55} & \cellcolor{tabsalmon!0}{64.23} & \cellcolor{tabsalmon!1}{68.59} & \cellcolor{tabsalmon!0}{69.32} & \cellcolor{tabsalmon!0}{66.62} & \cellcolor{tabsalmon!0}{68.09} & \cellcolor{tabsalmon!0}{62.03} \\
\addlinespace[3pt]
 & \multirow{2}{*}{\textsc{Llama}\textsubscript{Human transcript}} & Std. & \cellcolor{tabsalmon!18}{58.45} & \cellcolor{tabsalmon!7}{65.45} & \cellcolor{tabsalmon!16}{69.09} & \cellcolor{tabsalmon!21}{72.33} & \cellcolor{tabsalmon!17}{73.23} & \cellcolor{tabsalmon!2}{67.60} & \cellcolor{tabsalmon!16}{71.89} & \cellcolor{tabsalmon!14}{67.88} \\
 &  & Disfl. & \cellcolor{tabsalmon!10}{54.45} & \cellcolor{tabsalmon!0}{63.17} & \cellcolor{tabsalmon!0}{64.45} & \cellcolor{tabsalmon!0}{67.66} & \cellcolor{tabsalmon!6}{71.22} & \cellcolor{tabsalmon!3}{67.86} & \cellcolor{tabsalmon!3}{69.28} & \cellcolor{tabsalmon!0}{63.98} \\
\addlinespace[3pt]
 & \multirow{2}{*}{\textsc{Tower}\textsubscript{Canary ASR}} & Std. & \cellcolor{tabsalmon!9}{54.16} & \cellcolor{tabsalmon!30}{72.48} & \cellcolor{tabsalmon!26}{71.25} & \cellcolor{tabsalmon!26}{73.41} & \cellcolor{tabsalmon!25}{74.59} & \cellcolor{tabsalmon!29}{73.94} & \cellcolor{tabsalmon!29}{74.47} & \cellcolor{tabsalmon!28}{71.24} \\
 &  & Disfl. & \cellcolor{tabsalmon!0}{49.45} & \cellcolor{tabsalmon!26}{71.34} & \cellcolor{tabsalmon!18}{69.53} & \cellcolor{tabsalmon!24}{72.97} & \cellcolor{tabsalmon!19}{73.47} & \cellcolor{tabsalmon!25}{73.11} & \cellcolor{tabsalmon!26}{73.82} & \cellcolor{tabsalmon!26}{70.78} \\
 & \multirow{2}{*}{\textsc{Tower}\textsubscript{Whisper ASR}} & Std. & \cellcolor{tabsalmon!6}{52.74} & \cellcolor{tabsalmon!25}{70.91} & \cellcolor{tabsalmon!20}{69.89} & \cellcolor{tabsalmon!23}{72.75} & \cellcolor{tabsalmon!21}{73.99} & \cellcolor{tabsalmon!26}{73.19} & \cellcolor{tabsalmon!24}{73.45} & \cellcolor{tabsalmon!27}{71.05} \\
 &  & Disfl. & \cellcolor{tabsalmon!0}{49.07} & \cellcolor{tabsalmon!24}{70.53} & \cellcolor{tabsalmon!18}{69.48} & \cellcolor{tabsalmon!22}{72.71} & \cellcolor{tabsalmon!15}{72.75} & \cellcolor{tabsalmon!26}{73.30} & \cellcolor{tabsalmon!20}{72.62} & \cellcolor{tabsalmon!27}{70.92} \\
\addlinespace[3pt]
 & \multirow{2}{*}{\textsc{Tower}\textsubscript{Human transcript}} & Std. & \cellcolor{tabsalmon!4}{51.85} & \cellcolor{tabsalmon!38}{74.63} & \cellcolor{tabsalmon!40}{74.25} & \cellcolor{tabsalmon!38}{75.73} & \cellcolor{tabsalmon!41}{77.54} & \cellcolor{tabsalmon!35}{75.22} & \cellcolor{tabsalmon!45}{77.56} & \cellcolor{tabsalmon!35}{72.96} \\
 &  & Disfl. & \cellcolor{tabsalmon!1}{50.30} & \cellcolor{tabsalmon!34}{73.64} & \cellcolor{tabsalmon!37}{73.52} & \cellcolor{tabsalmon!33}{74.73} & \cellcolor{tabsalmon!36}{76.68} & \cellcolor{tabsalmon!34}{75.02} & \cellcolor{tabsalmon!31}{74.71} & \cellcolor{tabsalmon!34}{72.79} \\
\midrule
\multirow{2}{*}{End-to-end} & \textsc{OWSM} & N/A & \cellcolor{tabsalmon!0}{40.97} & \cellcolor{tabsalmon!0}{30.89} & \cellcolor{tabsalmon!0}{49.48} & \cellcolor{tabsalmon!0}{50.40} & \cellcolor{tabsalmon!0}{47.71} & -- & \cellcolor{tabsalmon!0}{47.19} & \cellcolor{tabsalmon!0}{56.03} \\
 & \textsc{Canary} & N/A & -- & \cellcolor{tabsalmon!25}{70.80} & \cellcolor{tabsalmon!13}{68.26} & \cellcolor{tabsalmon!18}{71.91} & \cellcolor{tabsalmon!20}{73.77} & -- & \cellcolor{tabsalmon!19}{72.35} & -- \\
\midrule
\multirow{4}{*}{SpeechLLM} & \multirow{2}{*}{\textsc{Phi-4-MM-instruct}} & Std. & \cellcolor{tabsalmon!0}{37.96} & \cellcolor{tabsalmon!0}{33.13} & \cellcolor{tabsalmon!10}{67.62} & \cellcolor{tabsalmon!2}{68.65} & \cellcolor{tabsalmon!0}{63.45} & \cellcolor{tabsalmon!0}{51.83} & \cellcolor{tabsalmon!1}{68.83} & \cellcolor{tabsalmon!7}{66.15} \\
 &  & Disfl. & \cellcolor{tabsalmon!0}{18.63} & \cellcolor{tabsalmon!0}{18.72} & \cellcolor{tabsalmon!0}{28.66} & \cellcolor{tabsalmon!0}{38.37} & \cellcolor{tabsalmon!0}{23.09} & \cellcolor{tabsalmon!0}{21.54} & \cellcolor{tabsalmon!0}{49.98} & \cellcolor{tabsalmon!0}{47.22} \\
 & \multirow{2}{*}{\textsc{Qwen2.5-Omni}} & Std. & \cellcolor{tabsalmon!0}{41.14} & \cellcolor{tabsalmon!0}{29.15} & \cellcolor{tabsalmon!0}{38.80} & \cellcolor{tabsalmon!0}{39.09} & \cellcolor{tabsalmon!0}{38.31} & \cellcolor{tabsalmon!0}{36.24} & \cellcolor{tabsalmon!0}{38.48} & \cellcolor{tabsalmon!0}{57.45} \\
 &  & Disfl. & \cellcolor{tabsalmon!0}{46.07} & \cellcolor{tabsalmon!0}{31.78} & \cellcolor{tabsalmon!0}{49.54} & \cellcolor{tabsalmon!0}{50.42} & \cellcolor{tabsalmon!0}{45.01} & \cellcolor{tabsalmon!0}{35.60} & \cellcolor{tabsalmon!0}{46.80} & \cellcolor{tabsalmon!0}{63.88} \\
\midrule
\multirow{12}{*}{Commercial} & \multirow{2}{*}{\textsc{Gemini}\textsubscript{Human transcript}} & Std. & \cellcolor{tabsalmon!42}{70.45} & \cellcolor{tabsalmon!33}{73.29} & \cellcolor{tabsalmon!29}{71.83} & \cellcolor{tabsalmon!30}{74.19} & \cellcolor{tabsalmon!29}{75.28} & \cellcolor{tabsalmon!34}{75.01} & \cellcolor{tabsalmon!30}{74.58} & \cellcolor{tabsalmon!31}{72.04} \\
 &  & Disfl. & \cellcolor{tabsalmon!39}{68.82} & \cellcolor{tabsalmon!29}{71.99} & \cellcolor{tabsalmon!23}{70.49} & \cellcolor{tabsalmon!19}{72.08} & \cellcolor{tabsalmon!22}{74.04} & \cellcolor{tabsalmon!29}{73.90} & \cellcolor{tabsalmon!25}{73.65} & \cellcolor{tabsalmon!31}{72.00} \\
 & \multirow{2}{*}{\textsc{ChatGPT}\textsubscript{Human transcript}} & Std. & \cellcolor{tabsalmon!46}{72.20} & \cellcolor{tabsalmon!40}{75.32} & \cellcolor{tabsalmon!35}{73.15} & \cellcolor{tabsalmon!37}{75.57} & \cellcolor{tabsalmon!37}{76.86} & \cellcolor{tabsalmon!40}{76.48} & \cellcolor{tabsalmon!40}{76.56} & \cellcolor{tabsalmon!41}{74.52} \\
 &  & Disfl. & \cellcolor{tabsalmon!37}{67.96} & \cellcolor{tabsalmon!30}{72.25} & \cellcolor{tabsalmon!27}{71.43} & \cellcolor{tabsalmon!22}{72.62} & \cellcolor{tabsalmon!26}{74.80} & \cellcolor{tabsalmon!32}{74.52} & \cellcolor{tabsalmon!24}{73.33} & \cellcolor{tabsalmon!35}{72.95} \\
\addlinespace[3pt]
 & \multirow{2}{*}{\textsc{Gemini}\textsubscript{ASR transcript}} & Std. & \cellcolor{tabsalmon!45}{71.96} & \cellcolor{tabsalmon!44}{76.67} & \cellcolor{tabsalmon!41}{74.38} & \cellcolor{tabsalmon!41}{76.33} & \cellcolor{tabsalmon!46}{78.55} & \cellcolor{tabsalmon!47}{78.22} & \cellcolor{tabsalmon!41}{76.87} & \cellcolor{tabsalmon!44}{75.15} \\
 &  & Disfl. & \cellcolor{tabsalmon!38}{68.52} & \cellcolor{tabsalmon!26}{71.32} & \cellcolor{tabsalmon!28}{71.63} & \cellcolor{tabsalmon!19}{72.12} & \cellcolor{tabsalmon!25}{74.67} & \cellcolor{tabsalmon!31}{74.31} & \cellcolor{tabsalmon!26}{73.83} & \cellcolor{tabsalmon!32}{72.33} \\
 & \multirow{2}{*}{\textsc{ChatGPT}\textsubscript{ASR transcript}} & Std. & \cellcolor{tabsalmon!50}{\textbf{74.16}} & \cellcolor{tabsalmon!50}{\textbf{78.32}} & \cellcolor{tabsalmon!50}{\textbf{76.44}} & \cellcolor{tabsalmon!50}{\textbf{78.08}} & \cellcolor{tabsalmon!50}{\textbf{79.21}} & \cellcolor{tabsalmon!50}{\textbf{78.81}} & \cellcolor{tabsalmon!50}{\textbf{78.59}} & \cellcolor{tabsalmon!50}{\textbf{76.69}} \\
 &  & Disfl. & \cellcolor{tabsalmon!37}{67.73} & \cellcolor{tabsalmon!20}{69.33} & \cellcolor{tabsalmon!19}{69.71} & \cellcolor{tabsalmon!0}{68.43} & \cellcolor{tabsalmon!4}{70.87} & \cellcolor{tabsalmon!28}{73.65} & \cellcolor{tabsalmon!5}{69.64} & \cellcolor{tabsalmon!31}{72.13} \\
\addlinespace[3pt]
 & \multirow{2}{*}{\textsc{Gemini}\textsubscript{Audio}} & Std. & \cellcolor{tabsalmon!30}{64.43} & \cellcolor{tabsalmon!11}{66.79} & \cellcolor{tabsalmon!3}{66.11} & \cellcolor{tabsalmon!7}{69.75} & \cellcolor{tabsalmon!11}{72.15} & \cellcolor{tabsalmon!14}{70.50} & \cellcolor{tabsalmon!6}{69.82} & \cellcolor{tabsalmon!2}{64.87} \\
 &  & Disfl. & \cellcolor{tabsalmon!26}{62.61} & \cellcolor{tabsalmon!0}{63.40} & \cellcolor{tabsalmon!0}{65.57} & \cellcolor{tabsalmon!0}{66.00} & \cellcolor{tabsalmon!0}{68.36} & \cellcolor{tabsalmon!13}{70.15} & \cellcolor{tabsalmon!0}{67.12} & \cellcolor{tabsalmon!10}{66.90} \\
 & \multirow{2}{*}{\textsc{ChatGPT}\textsubscript{Audio}} & Std. & \cellcolor{tabsalmon!0}{35.49} & \cellcolor{tabsalmon!0}{31.43} & \cellcolor{tabsalmon!0}{25.75} & \cellcolor{tabsalmon!0}{27.65} & \cellcolor{tabsalmon!0}{25.26} & \cellcolor{tabsalmon!0}{30.64} & \cellcolor{tabsalmon!0}{27.71} & \cellcolor{tabsalmon!0}{34.93} \\
 &  & Disfl. & \cellcolor{tabsalmon!0}{29.04} & \cellcolor{tabsalmon!0}{32.72} & \cellcolor{tabsalmon!0}{28.87} & \cellcolor{tabsalmon!0}{32.38} & \cellcolor{tabsalmon!0}{32.67} & \cellcolor{tabsalmon!0}{36.99} & \cellcolor{tabsalmon!0}{27.86} & \cellcolor{tabsalmon!0}{36.73} \\
\midrule
\multicolumn{11}{c}{\textit{X $\to$ EN Translations}} \\
\midrule
\multirow{2}{*}{Cascaded} & \textsc{Llama}\textsubscript{Human transcript} & Std. & \cellcolor{tabblue!0}{64.76} & \cellcolor{tabblue!0}{66.84} & \cellcolor{tabblue!0}{69.14} & \cellcolor{tabblue!0}{69.47} & \cellcolor{tabblue!0}{72.01} & \cellcolor{tabblue!0}{71.12} & \cellcolor{tabblue!0}{72.39} & \cellcolor{tabblue!0}{69.56} \\
 & \textsc{Tower}\textsubscript{Human transcript} & Std. & \cellcolor{tabblue!2}{67.23} & \cellcolor{tabblue!2}{69.14} & \cellcolor{tabblue!1}{69.88} & \cellcolor{tabblue!2}{72.14} & \cellcolor{tabblue!1}{72.87} & \cellcolor{tabblue!9}{76.60} & \cellcolor{tabblue!1}{73.30} & \cellcolor{tabblue!1}{70.36} \\
\midrule
\multirow{2}{*}{Commercial} & \textsc{ChatGPT}\textsubscript{Human transcript} & Std. & \cellcolor{tabblue!50}{\textbf{73.91}} & \cellcolor{tabblue!50}{\textbf{76.22}} & \cellcolor{tabblue!50}{\textbf{75.33}} & \cellcolor{tabblue!50}{\textbf{77.50}} & \cellcolor{tabblue!50}{\textbf{76.76}} & \cellcolor{tabblue!50}{\textbf{79.01}} & \cellcolor{tabblue!50}{\textbf{77.83}} & \cellcolor{tabblue!50}{\textbf{73.75}} \\
 & \textsc{Gemini}\textsubscript{Human transcript} & Std. & \cellcolor{tabblue!48}{73.66} & \cellcolor{tabblue!35}{74.07} & \cellcolor{tabblue!24}{72.45} & \cellcolor{tabblue!31}{75.32} & \cellcolor{tabblue!24}{74.66} & \cellcolor{tabblue!47}{78.82} & \cellcolor{tabblue!26}{75.57} & \cellcolor{tabblue!50}{\textbf{73.75}} \\
\bottomrule
\end{tabular}%
}
\caption{COMET-Kiwi scores across models, prompts, and target languages for English$\to$X and X$\to$English translation. Std.\ = standard prompt; Disfl.\ = disfluency-aware prompt.}
\label{tab:comet}
\end{table*}
\begin{table*}[t]
\centering
\begin{tabular}{llrrrrrrrr}
\toprule
 & & \multicolumn{2}{c}{\textbf{WER (\%)}} & \multicolumn{3}{c}{\textbf{E-Score}} & \multicolumn{3}{c}{\textbf{Z-Score}} \\
\textbf{Model} & \textbf{Prompt} & \textbf{Disfl. ref} & \textbf{Fluent ref} & \textbf{P} & \textbf{R} & \textbf{F1} & \textbf{EDITED} & \textbf{INTJ} & \textbf{PRN} \\
\midrule
\textsc{Whisper} & - & \cellcolor{tabsalmon!17}{20.31} & \cellcolor{tabsalmon!49}{22.27} & \cellcolor{tabsalmon!0}{68.69} & \cellcolor{tabsalmon!0}{67.47} & \cellcolor{tabsalmon!0}{64.91} & \cellcolor{tabsalmon!1}{86.22} & \cellcolor{tabsalmon!0}{71.20} & \cellcolor{tabsalmon!3}{32.33} \\
\textsc{Canary} & - & \cellcolor{tabsalmon!50}{15.69} & \cellcolor{tabsalmon!40}{24.99} & \cellcolor{tabsalmon!0}{64.33} & \cellcolor{tabsalmon!12}{55.03} & \cellcolor{tabsalmon!13}{54.53} & \cellcolor{tabsalmon!16}{68.02} & \cellcolor{tabsalmon!6}{63.99} & \cellcolor{tabsalmon!13}{26.43} \\
\textsc{ChatGPT} & Std. & \cellcolor{tabsalmon!0}{31.68} & \cellcolor{tabsalmon!43}{24.12} & \cellcolor{tabsalmon!7}{61.25} & \cellcolor{tabsalmon!0}{82.63} & \cellcolor{tabsalmon!0}{65.61} & \cellcolor{tabsalmon!0}{91.61} & \cellcolor{tabsalmon!0}{80.12} & \cellcolor{tabsalmon!0}{68.83} \\
\textsc{ChatGPT} & Disfl. & \cellcolor{tabsalmon!0}{22.93} & \cellcolor{tabsalmon!50}{22.01} & \cellcolor{tabsalmon!0}{64.44} & \cellcolor{tabsalmon!0}{67.42} & \cellcolor{tabsalmon!1}{63.00} & \cellcolor{tabsalmon!0}{88.44} & \cellcolor{tabsalmon!0}{70.46} & \cellcolor{tabsalmon!0}{36.28} \\
\textsc{Gemini} & Std. & \cellcolor{tabsalmon!10}{21.19} & \cellcolor{tabsalmon!0}{47.84} & \cellcolor{tabsalmon!50}{42.34} & \cellcolor{tabsalmon!48}{17.98} & \cellcolor{tabsalmon!50}{27.85} & \cellcolor{tabsalmon!49}{27.74} & \cellcolor{tabsalmon!48}{18.39} & \cellcolor{tabsalmon!48}{5.80} \\
\textsc{Gemini} & Disfl. & \cellcolor{tabsalmon!2}{22.35} & \cellcolor{tabsalmon!0}{51.36} & \cellcolor{tabsalmon!49}{42.97} & \cellcolor{tabsalmon!50}{16.45} & \cellcolor{tabsalmon!49}{28.73} & \cellcolor{tabsalmon!50}{26.56} & \cellcolor{tabsalmon!50}{16.11} & \cellcolor{tabsalmon!50}{4.39} \\
\bottomrule
\end{tabular}
\caption{Automatic speech recognition (ASR) results: word error rate (WER) against disfluent and fluent references, and disfluency retention (E-Score, Z-Score) for EDITED, INTJ, and PRN node types.}
\label{tab:asr_results}
\end{table*}
\begin{table*}[t]
\centering

\begin{tabular}{llrrrrrrrr}
\toprule
\textbf{Model} & \textbf{Prompt} & \textbf{AR} & \textbf{CS} & \textbf{DE} & \textbf{ES} & \textbf{FR} & \textbf{HI} & \textbf{IT} & \textbf{ZH} \\
\midrule
\multicolumn{10}{c}{\textit{Meaning}} \\
\midrule
\textsc{Tower}\textsubscript{Human transcript} & Std. & \cellcolor{tabsalmon!0}{47.19} & \cellcolor{tabsalmon!50}{83.69} & \cellcolor{tabsalmon!50}{83.48} & \cellcolor{tabsalmon!50}{85.43} & \cellcolor{tabsalmon!50}{85.28} & \cellcolor{tabsalmon!50}{82.79} & \cellcolor{tabsalmon!50}{84.58} & \cellcolor{tabsalmon!50}{81.01} \\
\textsc{Canary} & Std. & -- & \cellcolor{tabsalmon!23}{80.06} & \cellcolor{tabsalmon!1}{74.95} & \cellcolor{tabsalmon!2}{78.61} & \cellcolor{tabsalmon!9}{78.42} & -- & \cellcolor{tabsalmon!3}{76.58} & -- \\
\textsc{Phi-4-MM-instruct} & Std. & \cellcolor{tabsalmon!4}{53.08} & \cellcolor{tabsalmon!0}{49.99} & \cellcolor{tabsalmon!0}{72.67} & \cellcolor{tabsalmon!0}{75.38} & \cellcolor{tabsalmon!0}{63.68} & \cellcolor{tabsalmon!0}{63.12} & \cellcolor{tabsalmon!0}{71.96} & \cellcolor{tabsalmon!0}{68.06} \\
\textsc{Gemini}\textsubscript{Audio} & Std. & \cellcolor{tabsalmon!50}{78.94} & \cellcolor{tabsalmon!35}{81.66} & \cellcolor{tabsalmon!24}{78.96} & \cellcolor{tabsalmon!26}{82.03} & \cellcolor{tabsalmon!33}{82.49} & \cellcolor{tabsalmon!40}{81.04} & \cellcolor{tabsalmon!27}{80.73} & \cellcolor{tabsalmon!21}{76.44} \\
\midrule
\multicolumn{10}{c}{\textit{Style}} \\
\midrule
\textsc{Tower}\textsubscript{Human transcript} & Std. & \cellcolor{tabsalmon!0}{43.19} & \cellcolor{tabsalmon!47}{70.68} & \cellcolor{tabsalmon!50}{73.21} & \cellcolor{tabsalmon!50}{78.45} & \cellcolor{tabsalmon!50}{76.58} & \cellcolor{tabsalmon!50}{77.81} & \cellcolor{tabsalmon!50}{77.29} & \cellcolor{tabsalmon!50}{72.63} \\
\textsc{Canary} & Std. & -- & \cellcolor{tabsalmon!0}{37.31} & \cellcolor{tabsalmon!0}{41.83} & \cellcolor{tabsalmon!0}{39.51} & \cellcolor{tabsalmon!0}{43.26} & -- & \cellcolor{tabsalmon!0}{42.41} & -- \\
\textsc{Phi-4-MM-instruct} & Std. & \cellcolor{tabsalmon!3}{45.12} & \cellcolor{tabsalmon!4}{52.50} & \cellcolor{tabsalmon!7}{61.08} & \cellcolor{tabsalmon!7}{63.45} & \cellcolor{tabsalmon!3}{55.43} & \cellcolor{tabsalmon!2}{56.30} & \cellcolor{tabsalmon!6}{62.30} & \cellcolor{tabsalmon!0}{53.89} \\
\textsc{Gemini}\textsubscript{Audio} & Std. & \cellcolor{tabsalmon!50}{56.64} & \cellcolor{tabsalmon!50}{71.91} & \cellcolor{tabsalmon!44}{71.61} & \cellcolor{tabsalmon!31}{71.75} & \cellcolor{tabsalmon!48}{75.75} & \cellcolor{tabsalmon!0}{54.43} & \cellcolor{tabsalmon!35}{72.13} & \cellcolor{tabsalmon!20}{65.50} \\
\bottomrule
\end{tabular}
\caption{LLM-judge Meaning and Style Preservation scores by system and language.}
\label{tab:llm_judge}
\end{table*}
\begin{table*}[t]
\centering

\begin{tabular}{llcrrrrrrrr}
\toprule
\textbf{Metric} & \textbf{Ref?} & \textbf{Human Axis} & \textbf{AR} & \textbf{CS} & \textbf{DE} & \textbf{ES} & \textbf{FR} & \textbf{HI} & \textbf{IT} & \textbf{ZH} \\
\midrule
\multicolumn{11}{c}{\textit{Segment-Corr}} \\
\midrule
\mbox{LLM-judge}\textsubscript{Style} & \multirow{3}{*}{$\times$} & Style & 33.4 & -4.5 & 37.9 & 51.4 & \textbf{45.7} & 43.6 & 44.7 & 46.0 \\
\mbox{LLM-judge}\textsubscript{Meaning} &  & Meaning & 56.6 & 0.1 & 29.3 & 32.3 & 39.7 & 36.9 & 30.4 & 48.0 \\
\mbox{LLM-judge}\textsubscript{Overall} &  & Overall & 47.1 & -1.5 & 38.8 & 52.4 & \textbf{48.1} & 47.2 & 44.7 & 59.8 \\
\cmidrule{3-11}
\mbox{LLM-judge}\textsubscript{Style} & \multirow{3}{*}{$\checkmark$} & Style & 38.1 & 0.8 & 38.7 & \textbf{53.2} & 44.1 & \textbf{53.7} & \textbf{49.5} & 45.4 \\
\mbox{LLM-judge}\textsubscript{Meaning} &  & Meaning & \textbf{60.5} & 5.4 & \textbf{38.3} & \textbf{44.7} & \textbf{39.7} & 53.2 & 32.8 & \textbf{51.5} \\
\mbox{LLM-judge}\textsubscript{Overall} &  & Overall & 49.4 & 3.4 & \textbf{43.0} & \textbf{55.0} & 45.4 & \textbf{63.4} & \textbf{52.6} & \textbf{62.6} \\
\cmidrule{3-11}
\multirow{3}{*}{chrF} & \multirow{3}{*}{$\checkmark$} & Style & \textbf{45.8} & 8.6 & \textbf{38.9} & 38.9 & 37.6 & 52.2 & 37.5 & \textbf{46.1} \\
 &  & Meaning & 53.9 & \textbf{9.0} & 28.4 & 36.0 & 26.8 & \textbf{56.9} & 22.0 & 43.5 \\
 &  & Overall & \textbf{51.2} & 10.6 & 36.4 & 40.9 & 34.4 & 55.9 & 30.4 & 49.1 \\
\cmidrule{3-11}
\multirow{3}{*}{\mbox{COMET-Kiwi}} & \multirow{3}{*}{$\times$} & Style & 40.4 & \textbf{9.0} & 31.2 & 7.9 & 22.2 & 45.5 & 23.5 & 35.4 \\
 &  & Meaning & 47.5 & 8.4 & 32.2 & 18.7 & 28.9 & 48.2 & \textbf{37.5} & 45.7 \\
 &  & Overall & 47.2 & \textbf{11.0} & 32.6 & 12.2 & 26.3 & 44.8 & 31.2 & 42.0 \\
\midrule
\multicolumn{11}{c}{\textit{System-Corr}} \\
\midrule
\mbox{LLM-judge}\textsubscript{Style} & \multirow{3}{*}{$\times$} & Style & 92.0 & 61.5 & 93.0 & \textbf{99.4} & 80.8 & 77.1 & 99.5 & 90.8 \\
\mbox{LLM-judge}\textsubscript{Meaning} &  & Meaning & 70.9 & 68.8 & \textbf{97.5} & 93.5 & \textbf{99.2} & 96.5 & \textbf{99.4} & 95.8 \\
\mbox{LLM-judge}\textsubscript{Overall} &  & Overall & 78.1 & 72.2 & 95.8 & \textbf{99.9} & 93.6 & 99.8 & \textbf{93.9} & 94.8 \\
\cmidrule{3-11}
\mbox{LLM-judge}\textsubscript{Style} & \multirow{3}{*}{$\checkmark$} & Style & \textbf{99.4} & 68.8 & \textbf{96.7} & 99.4 & 86.5 & 98.8 & \textbf{99.8} & 91.0 \\
\mbox{LLM-judge}\textsubscript{Meaning} &  & Meaning & 74.0 & 67.2 & 92.0 & \textbf{96.2} & 99.2 & \textbf{99.9} & 97.4 & 96.1 \\
\mbox{LLM-judge}\textsubscript{Overall} &  & Overall & 84.1 & 75.3 & \textbf{96.0} & \textbf{99.9} & 96.6 & \textbf{100.0} & \textbf{93.9} & 94.8 \\
\cmidrule{3-11}
\multirow{3}{*}{chrF} & \multirow{3}{*}{$\checkmark$} & Style & 85.2 & 84.4 & 91.3 & 99.0 & \textbf{96.4} & \textbf{100.0} & 97.1 & \textbf{91.1} \\
 &  & Meaning & \textbf{98.1} & 68.0 & 74.4 & 90.2 & 99.1 & \textbf{99.9} & 81.0 & \textbf{96.3} \\
 &  & Overall & \textbf{94.0} & 75.8 & 87.6 & 99.5 & \textbf{99.8} & \textbf{100.0} & 91.2 & \textbf{95.0} \\
\cmidrule{3-11}
\multirow{3}{*}{\mbox{COMET-Kiwi}} & \multirow{3}{*}{$\times$} & Style & 85.2 & \textbf{85.4} & 62.7 & 63.3 & 81.2 & \textbf{100.0} & 63.0 & 84.5 \\
 &  & Meaning & \textbf{98.1} & \textbf{83.1} & 71.9 & 73.3 & 85.3 & \textbf{99.9} & 79.1 & 79.3 \\
 &  & Overall & \textbf{94.0} & \textbf{87.1} & 64.1 & 63.9 & 84.7 & \textbf{100.0} & 68.9 & 80.7 \\
\bottomrule
\end{tabular}%

\caption{Correlation (\%) between human Style/Meaning/Overall judgments and automatic metrics, averaged over the eight languages and four systems in the human evaluation. Segment-level: mean Kendall's $\tau$ (within-item, across systems). System-level: soft pairwise accuracy (SPA).}
\label{tab:correlation_per_language}
\end{table*}


\iftaclpubformat
\fi
\end{document}